\documentclass[conference]{IEEEtran}
\IEEEoverridecommandlockouts

\usepackage{authblk}
\usepackage{hyperref}
\usepackage{setspace}
\usepackage{amsmath }
\usepackage{times}
\usepackage{mathptmx}
\usepackage{verbatim}
\usepackage{adjustbox}
\usepackage{subfig}
\usepackage{graphicx}
\usepackage{color}
\usepackage{url}
\usepackage{hyperref}
\usepackage{listings}
\usepackage[normalem]{ulem}
\usepackage{setspace}
\usepackage{algorithm}
\usepackage{algorithmicx}
\usepackage[noend]{algpseudocode}
\usepackage[normalem]{ulem}
\usepackage{amssymb}
\usepackage{xparse}
\usepackage{xspace}
\usepackage{mathtools}
\usepackage{amsmath}
\usepackage{multirow}
\usepackage{xcolor}
\usepackage{rotating} 

\DeclareMathOperator*{\argminA}{arg\,min}
\DeclareMathOperator*{\avgA}{Av\,g}

\usepackage{eqparbox} 
\newdimen{\algindent}
\algnewcommand\LeftComment[2]{%
\hspace{#1\algindent}$\triangleright$ #2 \hfill %
}

\def\BibTeX{{\rm B\kern-.05em{\sc i\kern-.025em b}\kern-.08em
    T\kern-.1667em\lower.7ex\hbox{E}\kern-.125emX}}

\usepackage{caption}
\newcommand{\distance}{0pt}
\newcommand{\tech}{DeepBillboard\xspace}
\newcommand{\remove}[1]{}

\begin{document}
\bstctlcite{IEEEexample:BSTcontrol}

\title{\tech: Systematic Physical-World Testing of Autonomous Driving Systems} 

\author[1]{Husheng Zhou}
\author[2,3]{Wei Li}
\author[1]{Yuankun Zhu}
\author[2]{Yuqun Zhang}
\author[3]{Bei Yu}
\author[1]{Lingming Zhang}
\author[1]{Cong Liu}
\affil[1]{The University of Texas at Dallas}
\affil[2]{Southern University of Science and Technology}
\affil[3]{The Chinese University of Hong Kong}

\maketitle
\thispagestyle{plain}
\pagestyle{plain}

\begin{abstract} Deep Neural Networks (DNNs) have been widely applied in many autonomous systems such as autonomous driving and robotics for their state-of-the-art, even human-competitive accuracy in cognitive computing tasks. Recently, DNN testing has been intensively studied to automatically generate adversarial examples, which inject small-magnitude perturbations into inputs to test DNNs under extreme situations. While existing testing techniques prove to be effective, particularly for autonomous driving, they mostly focus on generating \emph{digital} adversarial perturbations, e.g., changing image pixels, which may never happen in physical world. Thus, there is a critical missing piece in the literature on autonomous driving testing: understanding and exploiting both \emph{digital} and \emph{physical} adversarial perturbation generation for impacting steering decisions. In this paper, we propose a systematic physical-world testing approach, namely \tech, targeting at a quite common and practical driving scenario: drive-by billboards. \tech is capable of generating a robust and resilient printable adversarial billboard test, which works under dynamic changing driving conditions including viewing angle, distance, and lighting. The objective is to maximize the possibility, degree, and duration of the steering-angle errors of an autonomous vehicle driving by our generated billboard with adversarial perturbations\remove{ of an autonomous vehicle driving by our generated roadside billboard with adversarial perturbations (applied both digitally and physically)}. We have extensively evaluated the efficacy and robustness of \tech~through conducting both experiments with digital perturbations and physical-world case studies. The digital experimental results show that \tech~is effective for various steering models and scenes. 
Furthermore, the physical case studies demonstrate that \tech~is sufficiently robust and resilient for generating physical-world adversarial billboard tests for real-world driving under various weather conditions, being able to mislead the average steering angle error up to 26.44 degrees. To the best of our knowledge, this is the first study demonstrating the possibility of generating realistic and continuous physical-world tests for practical autonomous driving systems; moreover, the basic \tech approach can be directly generalized to a variety of other physical entities/surfaces along the curbside, e.g., a graffiti painted on a wall.
\end{abstract}

\section{Introduction}

Deep Neural Networks (DNNs) are being widely applied in many autonomous systems for their state-of-the-art, even human-competitive accuracy in cognitive computing tasks. One such domain is autonomous driving, where DNNs are used to map the raw pixels from on-vehicle cameras to the steering control decisions~\cite{chen2015deepdriving, px2}. Recent end-to-end learning frameworks make it even possible for DNNs to learn to self-steer from limited human driving datasets~\cite{dave}.

Unfortunately, the reliability and correctness of systems adopting DNNs as part of their control pipeline have not been formally guaranteed. In practice, such systems often misbehave in unexpected or incorrect manners, particularly in certain corner cases due to various reasons such as overfitted/underfitted DNN models, biased training data, or incorrect runtime parameters. Such misbehaviors may cause severe consequences given the safety-critical nature of autonomous driving.  
A recent example of tragedy is that an Uber self-driving car struck and killed an Arizona pedestrian because the autopilot system made an incorrect control decision that ``\emph{it didn't need to react right away}'' when the victim was crossing the road at night. Even worse, recent DNN testing research has shown that DNNs are rather vulnerable to intentional adversarial inputs with perturbations~\cite{carlini2016towards,kurakin2016adversarial,moosavi2016deepfool,papernot2016limitations,szegedy2013intriguing}. 
The root cause of adversarial inputs and how to systematically generate such inputs are being studied in many recent DNN testing works~\cite{eykholt2018robust,ccs16,eykholt2018robust,carlini2016towards,deepgauge,deepxplore,deeptest,deeproad}. While these works propose various testing techniques that prove to be effective, particularly for autonomous driving, they mainly focus on generating \emph{digital} adversarial perturbations, which may never happen in physical world. The only exception is a recent set of works~\cite{eykholt2018robust,ccs16}, which take first step in printing robust physical perturbations that lead to misclassification of static physical objects (i.e., printouts in~\cite{robuster}, human face in~\cite{ccs16}, and stop sign in~\cite{eykholt2018robust}). Our work seeks to further enhance physical-world testing of autonomous driving by enhancing test effectiveness during a realistic, continuous driving process. Focusing on generating adversarial perturbations on any single snapshot of any misclassified physical object is unlikely to work in practice, as any real-world driving scenario may encounter driving conditions (e.g., viewing angle/distance) that are dramatically different from those in that static single-snapshot view.

\begin{figure}[t]
\centering
\includegraphics[width=3.4in]{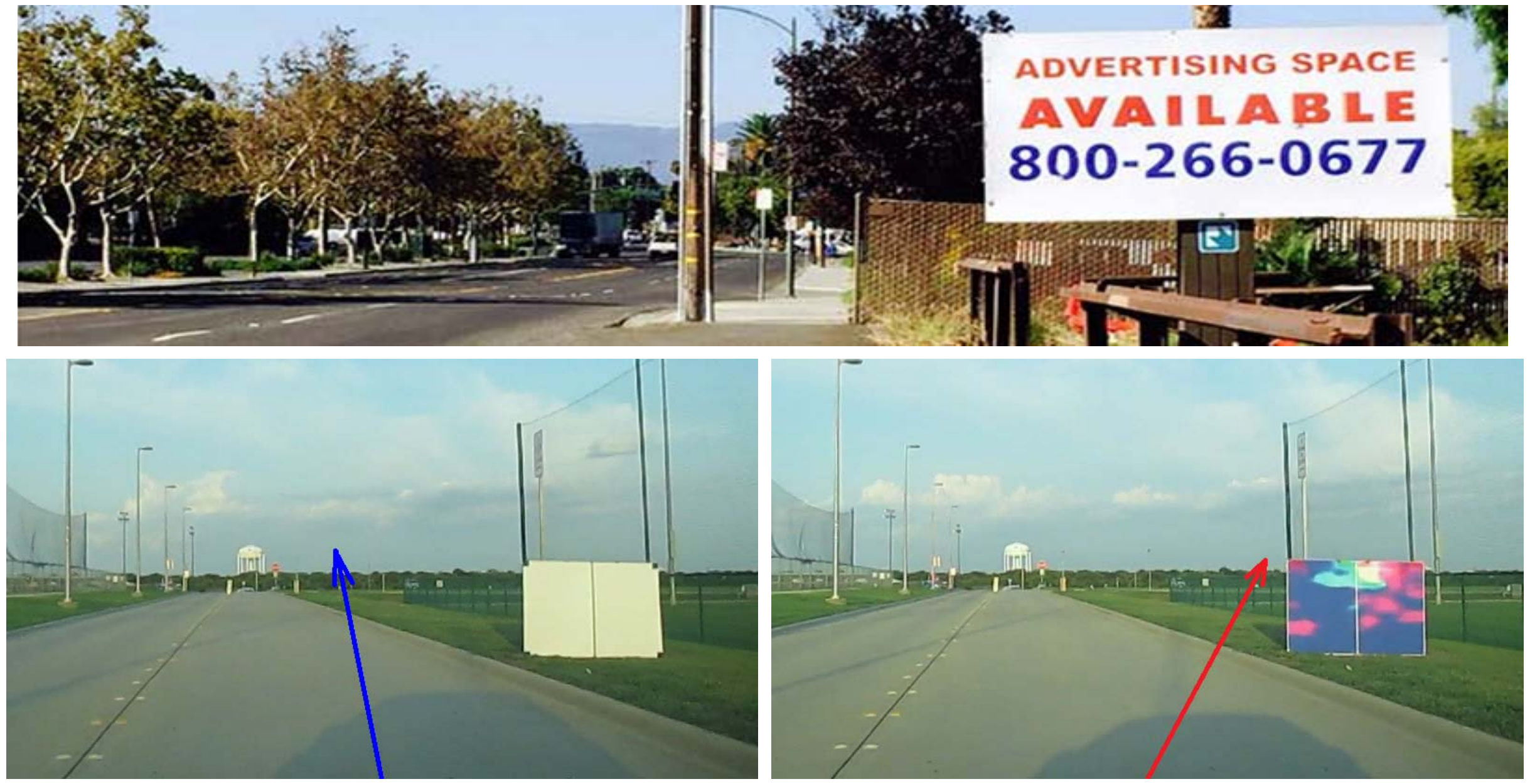}
\caption{The top subfigure shows an example customizable roadside billboard. 
The bottom two subfigures show an adversarial billboard example, where the Dave~\cite{dave} steering model diverges under our proposed approach.}
\label{fig:rent}
\end{figure}

In this paper, we propose a systematic physical-world testing approach, namely \tech, targeting at a quite common and practical continuous driving scenario: an autonomous vehicle drives by roadside billboards. \tech contributes to the systematic generation of adversarial examples for misleading steering angle when perturbations are added to roadside billboards in either a \emph{digital} or \emph{physical} manner.  Note that the basic idea can also be directly generalized to a variety of other physical entities/surfaces besides just billboards along the roadside, e.g., a graffiti painted on a wall; 
in this work, we choose the roadside billboards as our targeted physical driving scenario for several practical considerations: (1) Billboards are available to rent for advertising everywhere. Attackers who rent billboards can customize their sizes and contents, as illustrated in Fig.~\ref{fig:rent}; (2) Billboards are usually considered irrelevant or benign to the safety of transportation, and there are no strict rules regulating the appearance of a billboard; (3) Billboards are usually large enough to read by drivers and thus dashcams for cars with different distances, viewing angles, and light conditions; (4) An attacker may easily construct a physical world billboard to affect the steering decision of driving-by autonomous vehicles without others noticing, e.g., the actual core adversarial painting can only be a part of the entire billboard while the other parts of the billboard can still look normal, e.g., some bottom text bar showing ``Art Museum This Saturday''. 

The objective of \tech is to generate a single adversarial billboard image that may mislead the steering angle of an autonomous vehicle upon every single frame captured by onboard dashcam during the process of driving by a billboard. 
To generate effective perturbations, a major challenge is to cover a set of image frames exhibiting different conditions, including distance to the billboard, viewing angle, and lighting. 
Simply applying existing DNN testing techniques~\cite{deepxplore,deeptest,deeproad} to generate digital perturbations upon any specific frame clearly does not work in this case, because a realistic driving scene may not incur any frame with same or similar conditions (e.g., inserting sky black holes as done in the recent award-winning DeepXplore work~\cite{deepxplore}). 
Besides, the effectiveness of single frame perturbation may be not effective, since a mis-steering upon a frame may be quickly corrected by the next frame. 

To resolve this critical challenge, we develop a robust and resilient joint optimization algorithm, which generates a printable billboard image with perturbations that may mislead the steering angle upon every single frame captured by the dashcam during the entire driving process. 
To maximize the adversarial effectiveness, we develop various techniques to minimize interferences among per-frame-perturbations, and design the algorithm towards achieving global optimality considering all frames. 
Moreover, by inputting videos that record the process of driving by a roadside billboard with different driving patterns (e.g., driving speed and route), our algorithm can be easily tuned to generate printable adversarial image that is robust and resilient considering various physical world constraints such as changing environmental conditions and pixel printability due to printer hardware constraints.

\vspace{2mm}
\noindent \textbf{Contributions.} Considering such a real-world driving scenario and developing a corresponding digital and physical adversarial test generation method yield obvious advantages in terms of test effectiveness: the possibility, degree, and duration of misled steering decisions of any driving-by vehicles due to the adversarial billboards can be reliably increased. 
Our key contributions are summarized as follow.
\begin{enumerate}
\item We propose a novel angle of testing autonomous driving systems in the physical world that can be easily deployed. 

\item We introduce a robust joint optimization method to systematically generate adversarial perturbations that can be patched on roadside billboards both \emph{digitally} and \emph{physically} to consistently mislead steering decisions of an autonomous vehicle driving by the billboard with different driving patterns.

\item We propose new evaluation metrics and methodology to measure the test effectiveness of perturbations for steering models in both digital and physical domains. 

\item We prove the robustness and effectiveness of \tech~through conducting extensive experiments with both digital perturbations and physical case studies. The digital experimental results show that \tech~is effective for various steering models and scenes, being able to mislead the average steering angle up to 41.93 degree under various scenarios. The physical case studies further demonstrate that \tech~is sufficiently robust and resilient for generating physical-world adversarial billboard tests for real-world driving under various weather conditions, being able to mislead the average steering angle error from 4.86 up to 26.44 degree. \emph{To the best of our knowledge, this is the first study demonstrating the possibility of generating realistic and continuous physical-world tests for practical autonomous driving scenarios.}
\end{enumerate}

\section{Background and Related Work}

\noindent \textbf{DNN in Autonomous Driving.}
An autonomous driving system captures surrounding environmental data via multiple sensors (e.g. camera, Radar, Lidar) as inputs, processes these data with DNNs and outputs control decisions (e.g. steering).  
In this paper, we mainly focus on the steering angle component with camera inputs and steering angle outputs, as adopted in NVIDIA Dave~\cite{dave}.

Convolutional Neural Network (CNN), which is efficient at analyzing visual imagery, is the most widely used DNN for steering angle decisions. 
Similar to regular neural networks, CNNs are composed of multiple layers and pass information through layers in a feed-forward way. Among all layers, the convolutional layer is a key component in CNNs, which performs convolution with kernels on the output of previous layers and sends the feature maps to successor layers. Different from another widely used DNN architecture -- Recurrent Neural Networks (RNNs) which is a kind of neural network with feedback connections, CNN-based steering model makes steering decisions based only on the currently captured image. In this paper, we focus on the testing of CNN steering models and leave RNN testing as future work. We nonetheless note that \tech~can be adapted to apply to RNN testing. Intuitively, this can be achieved by modifying the gradient calculation method according to RNN's specific characteristics.
 
\noindent \textbf{Digital Adversarial Examples.}
Recent research shows that deep neural network classifier can be tested and further fooled by adversarial examples~\cite{carlini2016towards,kurakin2016adversarial,moosavi2016deepfool,papernot2016limitations,szegedy2013intriguing}. Such testing can be performed in both black-box~\cite{papernot2016transferability, papernot2017practical} and white-box~\cite{carlini2016towards,kurakin2016adversarial,moosavi2016deepfool,papernot2016limitations,szegedy2013intriguing} settings.  
Goodfellow et al. proposed the fast gradient method that applies a first-order approximation of the loss function to construct adversarial samples~\cite{goodfellow6572explaining}. Optimization-based methods have also been proposed to create adversarial perturbations for targeted attacks~\cite{carlini2016towards,liu2016delving}. Meanwhile,
the recent DeepTest~\cite{deeptest} and DeepRoad~\cite{deeproad} techniques transform original images to generate adversarial images via simple affine/filter transformations or Generative Adversarial Networks (GANs)~\cite{gan}.
Overall, these methods contribute to understanding digital adversarial examples, and the generated adversarial examples may never exist in reality (e.g., the rainy driving scenes generated by DeepTest~\cite{deeptest} and DeepRoad~\cite{deeproad} are still far from real-world scenes). By contrast, our work examines physical perturbations on real objects (billboards) under dynamic conditions such as changing distances and view angles.

\noindent \textbf{Physical Adversarial Examples.} 
Kurakin et al. showed that adversarial examples, when photoed by a smartphone camera, can still lead to misclassification~\cite{kurakin2016adversarial}. 
Athalye et al. introduced an attacking algorithm to generate physical adversarial examples that are robust to a set of synthetic transformations~\cite{athalye2017synthesizing}. They further created 3D-printed replicas of perturbed objects~\cite{athalye2017synthesizing}. 
The main differences between aforementioned works and our work include:
(1) Previous works only use a set of synthetic transformations during optimization, which can miss subtle physical effects; while our work can sample from both synthetic transformations and various real-world physical conditions.
(2) Our work modifies real-world true-sized objects; 
and (3) Our work targets the testing of realistic and continuous driving scenarios.

Sharif et al. presented dodging and impersonation attacks for DNN-based face recognition systems by printing adversarial perturbations on the eyeglasses frames~\cite{ccs16}. Their work demonstrated successful physical attacks in relatively stable physical conditions with little variation in pose, distance/angle from the camera, and lighting. This contributes an interesting understanding of physical examples in stable environments. However, environmental conditions can vary widely in general and can contribute to reducing the effectiveness of perturbations. Therefore, we choose the inherently unconstrained environment of drive-by billboards classification. In our work, we explicitly design our perturbations to be effective in the presence of diverse and continuous physical-world conditions (particularly, large distances/angles and resolution changes).

Lu et al. performed experiments with physical adversarial examples of road sign images against detectors and show that current detectors cannot be attacked~\cite{lu2017no}. Several more recent works have demonstrated adversarial examples against detection/segmentation algorithms digitally~\cite{DBLP:journals/corr/XieWZZXY17, metzen2017universal, cisse2017houdini}. 
The most recent work for attacking autonomous driving systems are the works conducted by Eykholt and Evtimov et al. They showed that physical robust attacks can be constructed for road signs classifiers~\cite{eykholt2018robust}, and such attacks can be further extended to attack YOLO detectors~\cite{eykholt2017note}. Our work differs from such works due to the fact that: (1) we target attacking steering models by constructing printable perturbations on drive-by billboards, which can be anywhere and have much more impacts than road signs; (2) our proposed algorithm considers a sequence of contiguous frames captured by dashcams with gradually changing distances and viewing angles, and seeks to maximize the possibility and the degree of misleading the steering angles of an autonomous vehicle driving by our adversarial roadside billboard; and (3) we introduce a new joint optimization algorithm to efficiently generate such attacks both digitally and physically.

\section{Generating Adversarial Billboards}

\subsection{Adversarial Scenarios}

The goal of \tech~is to mislead the steering angle of an autonomous vehicle, causing off-tracking from the central of the lane by painting the adversarial perturbation on the billboard alongside the road. Our targeted DNNs are CNN-based steering models~\cite{dave, dave2, dave3, cg32, rambo}, without involving detection/segmentation algorithms. The steering model takes images captured by dashcam as inputs, and outputs steering angle decisions.

We use off-tracking distance to measure the test effectiveness (i.e., the strength of steering misleading), which has been applied in Nvidia's Dave~\cite{dave} system to trigger human interventions. Assume the vehicle's speed is $v$ m/s, the decision frequency of using DNN inference is $i$ second(s), the ground truth steering angle is $\alpha$, and the misleading steering angle is $\alpha'$, then the off-tracking distance is calculated by $v \cdot i \cdot \sin(\alpha' - \alpha)$. 
In potential physical world attack, the speed of the vehicle usually are not controllable by the tester/attacker. Thus we use \textit{steering angle error} which is the steering angle divergence between ground truth and misled steering to measure the test effectiveness.

Instead of misleading the steering decision only at a fixed distance and view angle, which may be hardly captured by a driving-by vehicle, we consider the actual driving-by scenario. Specifically, when a vehicle is driving towards the billboard, we seek to generate a physical adversarial billboard that may mislead the steering decision upon a sequence of dashcam-captured frames viewing from different distances and angles. The number of captured frames clearly depends on the FPS of the dashcam and the time used for the vehicle to drive from the starting position till physically passing the billboard. 
Considering such a real-world dynamic driving scenario yields obvious advantage in terms of attacking strength: the possibility and the degree of misled steering decisions of any driving-by vehicles due to the adversarial billboards can be reliably increased. We emphasize that this consideration also fundamentally differentiate the algorithmic design of \tech from applying simpler strategies such as random search, average/max value-pooling, different order etc. Applying such simpler methods would improve misleading angle for a single frame yet lowering the overall objective. After a few iterations, such methods hardly improve the objective.






\subsection{Evaluating Matrices}
\label{sec:metrics}

Our evaluating metrics aim to reflect the attacking \textbf{strength} and \textbf{possibility}. Vehicles may pass by our adversarial billboard with different speeds and slightly different angles, which may impact the number of image frames captured by the camera and the billboard layout among different frames. Assume \^{X}=\{$x_0$, $x_1$, $x_2$, ... , $x_n$\} denotes an exhaustive set of image frames possibly captured by a drive-by vehicle with any driving pattern (e.g., driving speed and route), then frames captured by any drive-by vehicle are clearly a subset $X \subseteq$ \^{X}. Our objective is to generate the physical printable billboard which can affect (almost) every frame in \^{X}, such that any subset $X$ corresponding to a potential real-world driving scenario may have a maximized chance to be affected. To meet this objective, we define two evaluating metrics denoted $M_0$, $M_1$ as follows.


$M_0$ measures the mean angle error (MAE) for every frame in \^{X}:
\begin{equation} 
M_0 = \avgA_{0< i < \lVert \hat{X} \lVert} (f(x_i')-f(x_i)), 
\end{equation}
 where $f(\cdot)$ denotes the prediction result of the targeted steering model, $x'$ denotes the perturbed frame. This metric measures the average strength of attacks to the frame super set. A larger $M_0$ intuitively would imply a higher chance and a larger error of misleading the steering angle during the process of driving by the billboard.
 
$M_1$ measures the percentage of frames in \^{X} whose angle error exceeds a predefined threshold, denoted by $\tau$. $\tau$ can be calculated based on the physical driving behavior. A formal definition of $M_1$ is given by:
\begin{equation}
M_1 = \frac{\Vert \{x_i|f(x_i')-f(x_i) > \tau, 0 < i < \Vert \hat{X} \Vert \} \Vert }{\Vert \hat{X} \Vert}.
\end{equation}
 For example, if we want to mislead a 40MPH autonomous vehicle by an off-track distance of one meter within a time interval of 0.2 seconds,\footnote{We note that an autonomous vehicle would likely not run classification on every frame due to performance constraints, but rather classify every j-th frame, and then perform simple majority voting.} then $\tau$ can be calculated as 16.24 according to the above equation. We mainly adopt $M_1$ as an evaluating metric for our physical-world case studies, as $M_1$ can clearly reflect the number of frames that incur unacceptable steering decisions (e.g., those that may cause accidents) given any reasonable predefined threshold according to safety stands in practice.

\subsection{Challenges}

Physical attacks on an object should be able to work under changing conditions and remain effective at fooling the classifier. We structure our discussion of these conditions using our targeted billboard classification. A subset of these conditions can also be applied to other types of physical learning systems such as drones and robots.

\vspace{1mm}
\noindent \textbf{Spatial Constraints.} Existing adversarial algorithms mostly focus on perturbing digital images and add adversarial perturbations to all parts of the image, including background imagery (e.g., sky). However, for a physical billboard, the attacker cannot manipulate the background imagery other than the billboard area. Furthermore, the attacker cannot assume that there exists a fixed background imagery as it will change depending on the distance and viewing angle of the dashcam of a drive-by vehicle.

\vspace{1mm}
\noindent \textbf{Physical Limits on Imperceptibility.} An attractive feature of existing adversarial learning algorithms is that their perturbations to a digital image are often small in magnitude such that the perturbations are almost imperceptible to a casual observer. However, when transferring such minimal perturbations to a real world physical image, we must ensure that a camera is able to perceive the perturbations. Therefore, there are physical constraints on perturbation imperceptibility, which is also dependent on the sensing hardware.

\vspace{1mm}
\noindent \textbf{Environmental Conditions.}  The distance and angle of a camera in a drive-by autonomous vehicle with respect to a billboard may consistently vary. The captured frames that are fed into a classifier are taken at different distances and viewing angles. Therefore, any perturbation that an attacker physically adds to a billboard must be able to survive under such dynamics. Other impactful environmental factors include changes in lighting/weather conditions and the presence of debris on the camera or on the billboard.

\vspace{1mm}
\noindent \textbf{Fabrication Error.} To physically print out an image with all constructed perturbations, all perturbation values must be valid colors that can be printed in the real world. Furthermore, even if a fabrication device, such as a printer, can produce certain colors, there may exist certain pixel mismatching errors.

\vspace{1mm}
\noindent \textbf{Context Sensitivity.} Every frame in \^{X} must be perturbed considering its context in order to maximize the overall attacking strength (maximizing $M_0$ for instance). Each perturbed frame can be mapped to a printable adversarial image with a certain view angle and distance. Each standalone frame has its own optimal perturbation. However, we need to consider all frames' context to generate an ultimate single printable adversarial image that is globally optimal w.r.t. all frames.

In order to physically attack deep learning classifiers, an attacker should account for the above physical world constraints, for otherwise the effectiveness of perturbations can be significantly weakened.

\subsection{The Design of \tech}
\label{sec:design}

We design \tech, which generates a single printable image that can be pasted on a roadside billboard by analyzing given driving videos where vehicles drive by a roadside billboard with different driving patterns, for continuously misleading the steering angle decision of any drive-by autonomous vehicle. 
\tech starts with generating perturbations for every frame $f_i$ of a given video without considering frame context and other physical conditions. We then describe how to update the algorithm to resolve the aforementioned physical world challenges. We finally describe the algorithmic pseudocode of \tech~ in detail. We note that it may not be practically possible to construct the exhaustive set of image frames (i.e. \^{X}),  possibly captured by a drive-by vehicle with any driving pattern (e.g., driving speed and route). Nonetheless, processing a larger number of driving videos will clearly strengthen the testing effectiveness of \tech due to a larger \^{X}, 
at the cost of increased time complexity. 

The single frame adversarial example generation searches for a perturbation $\sigma$ to be added to the input x such that the perturbed input $x'=x+\sigma$ can be predicted by the targeted DNN steering model $f(\cdot)$ as
\begin{equation*}
max \quad H(f(x+\delta), A_x), 
\end{equation*}
where $H$ is a chosen distance function and $A_x$ is the ground truth steering angle. To solve the above constrained optimization problem, we reformulate it in the Lagrangian-relaxed form similar to prior work~\cite{eykholt2018robust, ccs16}:
\begin{equation}
\label{eq:obj}
\argminA_{\delta} (-L(f(x+\delta), A_x)), 
\end{equation}
where $L$ is the loss function which measures the difference between the model's prediction and  ground truth $A_x$. The attacking scenario in this paper can be treated as inference dodging which aims to not being correctly inferred.

\begin{figure*}[t]
\centering
\includegraphics[width=5in]{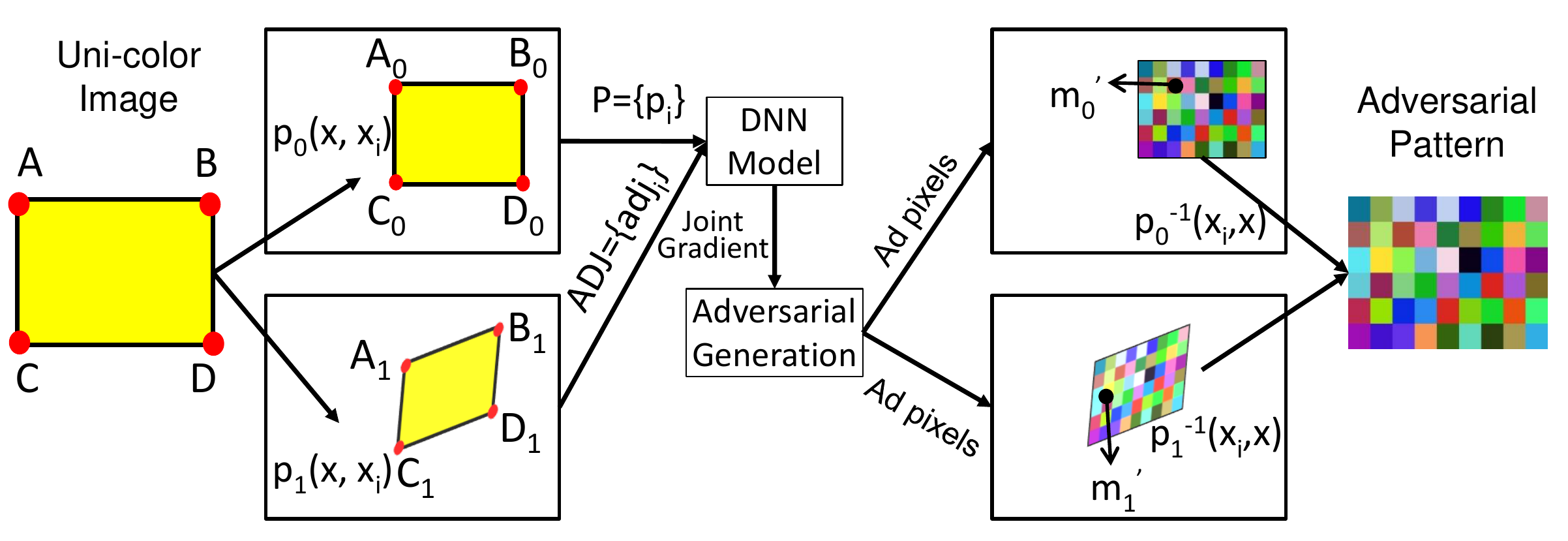}
\caption{Work flow of \tech to generate adversarial perturbations for contiguous frames.}
\label{fig:calib}
\end{figure*}

\vspace{2mm}
\noindent \textbf{Joint Loss Optimization.} As discussed earlier, our objective is to generate a single adversarial image that may mislead the steering angle of an autonomous vehicle upon every single frame the dashcam may capture during driving by the billboard.  
The appearance of the adversarial billboard may vary when being viewed from different angles and distances. As a result, to meet the objective, we need to generate one single printable adversarial perturbation that can mislead every single frame captured during the driving-by process.  
This is clearly an optimization problem beyond a single image. It is thus necessary to consider all frames jointly since one modification on the billboard affects all frames. To this end, the problem becomes finding a single perturbation $\Delta$ that optimizes Eq.~\ref{eq:obj} for every image $x$ in an image set $X$.
We formalize this perturbation generation as the following optimization problem.

\begin{equation}
\argminA_{\Delta} \sum_{0< i < \Vert X \Vert} (-L(f(x_i+p_i(\Delta)), A_x)), 
\end{equation}
where $p_i$ is the projection function of printable perturbation $\Delta$ into every single frame $i$. 

\vspace{2mm}
\noindent \textbf{Handling Overlapped Perturbations.} Every single frame may generate a set of perturbations which is composed of multiple pixels to be updated on the ultimate printable adversarial image. Perturbations of multiple frames may encounter overlapped pixels, which may produce interferences among those frames. 
To maximize the attacking strength, \tech~seeks to minimize the overlapped perturbations among multiple frames by only updating a fixed number of $k$ pixels for each single frame in order. The $k$ pixels are those that have the most impact on misleading the steering decision. 
We assume the final adversarial billboard image covering $n$ dashcam-captured frames is composed of $m$ pixels. $k$ is a value satisfying $n \cdot k < m$, which helps reduce the overall chance of perturbation overlapping among frames. For each overlapped pixel, we update it by greedily choosing a  value that maximizes the objective metric (e.g., $M_0$). 

\vspace{2mm}
\noindent \textbf{Enhancing Perturbation Printability.} For the perturbation to work in the physical world, each perturbed pixel needs to be a printable value by existing printer hardware. Let $P\subset [0,1]^3$ be the set of printable RGB triples. We define non-printability score (NPS) of a pixel to reflect the maximum distance between this pixel and any pixel in $P$. A larger NPS value would imply a smaller chance of accurately printing out the corresponding pixel. Our algorithm thus seeks to minimize NPS as part of the optimization. We define the NPS of a pixel $p'$ as:
\begin{equation}
NPS(p') =  \prod_{p \in P}|p'-p|.
\end{equation}
We  generalize the definition of NPS of a perturbation as the sum of NPS values of all the pixels in this perturbation. 



\vspace{2mm}
\noindent \textbf{Adjust Color Difference under Various Environment Conditions.} For different environmental conditions, the observable color of the same pixel belonging to the billboard image may look different in the video captured by a dashcam. Such difference may impact the adversarial efficacy under different conditions. In our physical world experiments, we pre-fill the entire billboard with unicolor $p=\{r,g,b\}$. Under a specific environment condition $e$, its actual color shown in camera may become $p'=\{r',g',b'\}$. Based on our experiments, we observe that such color differences of pixels in the same image are almost the same. To simplify the problem, we introduce a color adjustment function $ADJ_{i} = d_{i}(p, p')$ for each image $x_i$ to adjust the color difference. 

\begin{algorithm}[!t]
  \caption{Generating attacks for maximizing $M_0$}
  \label{alg:generate}
  \small{
	\begin{algorithmic}[1]
 	\Require IMGS \LeftComment{1}{List of images of the same scene}
 	\Require COORD \LeftComment{1}{List of coordinates of billboards in IMGS}
    \Require ITER \LeftComment{1}{Number of enhance iterations} 
    \Require BSZ \LeftComment{1}{Number of images in a batch} 
 	\Require ADJ \LeftComment{1}{List of adjustment for environment factors} 
 	\Require DIM \LeftComment{1}{Dimensions of printable perturbation}
	\Function{Generate}{}	
    \State perturb = COLOR\_INIT(DIM)
    \State pert\_data = zero(BSZ, DIM)
    \For {i in ITER}
    \State random.shuffle(IMGS)
    \For {j in range(0,len(IMGS), BSZ) }
    \State batch = IMGS[j, j+BSZ]
    \State pert\_data.clear()
    \For {x in batch}
    \State grad = $\partial obj / \partial x$
    \State grad = DOMAIN\_CONSTRNTS(grad)
    \State pert\_data[x] = REV\_PROJ(grad, ADJ)
    
    \EndFor
    
    \State pert\_data = HANDLE\_OVERLAP(pert\_data)
    \State atmpt\_pert = pert\_data $\cdot$ s + perturb
    \State atmpt\_pert = NPS\_CTL(atmp\_per, ADJ)
    \State atmpt\_imgs := UPDATE\_IMGS(atmpt\_pert, COORD)
    \State this\_diff = CALC\_DIFF(atmp\_imgs)
    \If {this\_diff $>$ last\_diff \textbf{or} rand() $<$ $SA$} 
    \State perturb = APPLY(perturb)
    \State imgs := UPDATE\_IMGS(perturb, COORD)
    \State last\_diff = this\_diff
    \EndIf
    \EndFor
    \EndFor
    \State \textbf{return} perturb
	\EndFunction
	\end{algorithmic}
    }
\end{algorithm}

\vspace{2mm}
\noindent \textbf{Algorithm overview.} The procedure of \tech~for generating an adversarial billboard image is illustrated in Fig.~\ref{fig:calib}. To generate an adversarial billboard image, we first pre-fill the billboard with unicolor, and paint its four corners with contrasting colors for the purpose of (1) locating the coordinates of the billboard digitally, and (2) getting the color adjustment function $ADJ_i$. Then we record video using dashcam and drive by the billboard with different driving behaviors (e.g., different driving speeds and driving patterns) along the road. Then we send the pre-recorded videos to our algorithm as inputs to generate the printable adversarial billboard image. As discussed earlier, inputting a larger number of driving videos will clearly strengthen the testing effectiveness of \tech, at the cost of increased time complexity.

The pseudocode of our adversarial algorithm is illustrated in Alg.~\ref{alg:generate}. 
Our algorithm is essentially iteration-based. In each iteration, we first obtain perturbation proposals for a batch of randomly chosen images according to their gradients which reflect the influence of every pixel to the final objective. We then greedily apply only those proposed perturbations that may lead to better adversarial effect. We apply a sufficient number of iterations to maximize steering angle divergence and the perturbation robustness.

As seen at the beginning of Alg.~\ref{alg:generate}, the inputs include: a list of frames in the pre-recorded videos, a list of coordinates of the four corners in the billboard in every frame, number of enhancing iterations, batch size, a list of color adjustment factors, and the dimension of targeted digital perturbation. 
As illustrated in Alg.~\ref{alg:generate}, $perturb$ is a printable perturbation matrix that is composed of RGB pixel values (line 2). We use $COLOR\_INIT$ to pre-fill the printable perturbation matrix with one unicolor $c \in \{0|255\}^3$. Based on our extensive digital and physical experiments, using unicolor-prefilled matrix may result in better results and faster convergence. According to our experiments, gold, blue, and green are the most efficient unicolors for our testing purposes. $pert\_data$ is a list of matrices which store the proposed perturbations for a batch of images (line 2). 
Lines 4 to 21 loop through enhanced iterations which aim to maximize the adversarial effectiveness and the perturbation robustness. 
At line 5, we randomly shuffle the processing order of captured frames. The purpose is to avoid quick convergence to a non-optimal point at early video frames (similar to deep neural network training). Starting from line 6, we loop over all the images which are split into batches. 
For each image batch, we initialize and clear $pert\_data$ (lines 7-8) before looping over every single image inside the batch (line 9). 

For each image $x$ within a batch, we calculate its gradient which is the partial derivative~\cite{goodfellow6572explaining} of object function to input image (line 10). By iteratively changing $x$ using gradient ascent, the object function can be easily maximized. We note that we can intentionally mislead targeted steering model to steer left or right by selecting positive or negative value of gradient.
We then apply domain constraints to the gradient (line 11) to ensure that we only update the pixels belonging to the corresponding area of the billboard, and the pixel values after gradient ascent are within a certain range (e.g., 0 to 255).
In the implementation, as discussed earlier, we introduce a parameter $k$ to only apply top $k$ gradient values that have the most impact on adversarial efficacy. This is to reduce the overlapped perturbations among all images.   
Different from the saliency map used in JSMA~\cite{jsma} which represents the confidence score of $x$ being classified into targeted class for the current image, we consider the influence to the joint objective function for all images in this scene, seeking to maximize the average steering angle difference from ground truth. 
After constraining the applicable gradients, we project the gradient values for each image $x$ to the proposed perturbations of the batched images (line 12). ADJ (i.e., the input list of adjustments for environment factors) is used to correct color difference for different lighting conditions. For example, if a pure yellow color $(255,255,0)$ becomes $(200,200,0)$, then $ADJ$ is set to be $(55,55,0)$. When projected to the physical billboard, the gradient value should be increased by $(55,55,0)$.  
  
After all images in the batch get their gradients, there may exist overlapped perturbations among them. That is, for each pixel corresponding to overlapped perturbations, it may have multiple proposed update values for the ultimate printable adversarial example. 
To handle such overlaps (line 13), we implemented three methods: (1) update the overlapped pixels with the max gradient value among proposed perturbations, (2) update the overlapped pixels with the sum of all gradient values, and (3) update the overlapped pixels with one of the proposed values that has the greatest overall influence to the objective function.  
 Then at line 14, we calculate the proposed update $atmpt\_pert$ by adding gradients to the current physical perturbation $perturb$.
After color corrections and non-printable score control (line 15), the proposed perturbations for the physical billboard are projected to the images according to the coordinates (line 16).
We calculate the total steering angle difference for perturbed images (line 17).
If the proposed perturbations can improve the objective, or meet the simulated annealing~\cite{van1987simulated}, indicated by $SA$.
we accept the proposed perturbations (line 19) and update all images with these perturbations (line 20). 
Then we record the current iteration's total steering angle divergence and use it as the starting point in the next iteration. 
When all enhanced iterations are finished, we return the physical perturbation $perturb$ as the resultant output. We note that, although our major goal is to generate physical perturbations, the output can be directly patched to digital images as well.

\vspace{-2mm}
\section{Evaluation}
\label{sec:eval}

In this section, we evaluate the efficacy of \tech~both digitally and physically for various steering models and road scenes.

\subsection{Experiment Setup}
\label{sec:setup}

\begin{table}[t]
\centering
\caption{Studied scenes for digital experiments.}
\label{tb:scenes}
\small{
\begin{tabular}{|l|l|l|l|l|}
\hline
\multicolumn{1}{|c|}{Scenes}  & \multicolumn{1}{c|}{Img} & \multicolumn{1}{c|}{Size} & \multicolumn{1}{c|}{BB\_min} & \multicolumn{1}{c|}{BB\_max} \\ \hline\hline
Dave-straight1     & 54  & $455 \times 256$ & $21 \times 22$ & $41 \times 49$  \\ \hline
Dave-curve1  & 34  & $455 \times 256$ & $29 \times 32$ & $51 \times 49$  \\ \hline
Udacity-straight1    & 22  & $640 \times 480$ & $48 \times 29$ & $66 \times 35$  \\ \hline
Udacity-curve1  & 80  & $640 \times 480$ & $51 \times 51$ & $155 \times 156$  \\ \hline
Kitti-straight1  & 20  & $455 \times 1392$ & $56 \times 74$ & $121 \times 162$  \\ \hline
Kitti-straight2    & 21  & $455 \times 1392$ & $80 \times 46$ & $247 \times 100$  \\ \hline
Kitti-curve1     & 21  & $455 \times 1392$ & $64 \times 74$ & $173 \times 223$  \\ \hline
\end{tabular}
}
\end{table}

\newcommand{\figw}{.115\textwidth}
\newcommand{\figh}{2.8cm}

\begin{sidewaystable*}
\caption{Average steering angle errors for various scenes. (The table is rotated clockwise 90 degree specifically for  better presentation clarity.)}
\label{tb:overall}
\centering
\begin{tabular}{|c|c|c|c|c|c|c|c|}
\hline
\multirow{2}{*}{Model} & \multicolumn{2}{c|}{Dave} & \multicolumn{2}{c|}{Udacity} & \multicolumn{3}{c|}{Kitti} \\ \cline{2-8} 
& Straight1      & Curve1      & Straight1        & Curve1       & Straight1  & Straight2  & Curve1 \\ \hline
Dave\_V1	
& \begin{minipage}{\figw} 
\includegraphics[width=\textwidth,height=\figh]{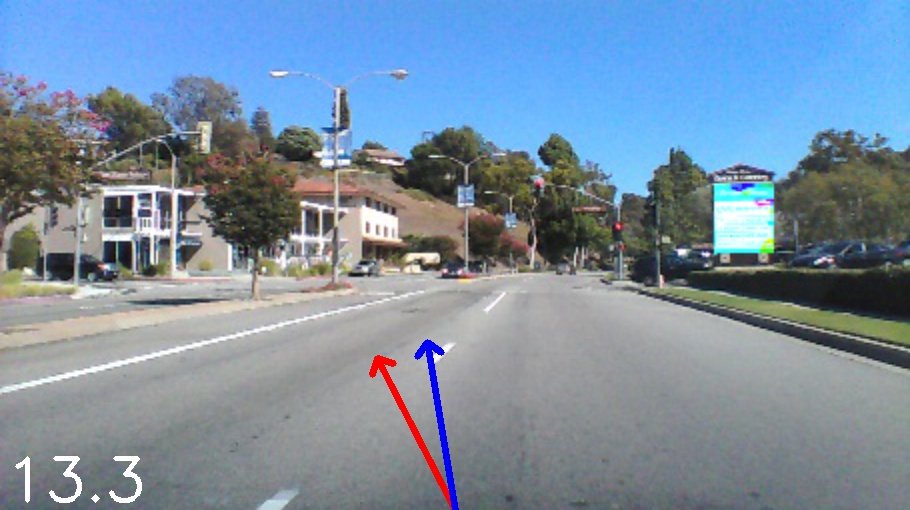} \end{minipage}         
& \begin{minipage}{\figw} 
\includegraphics[width=\textwidth,height=\figh]{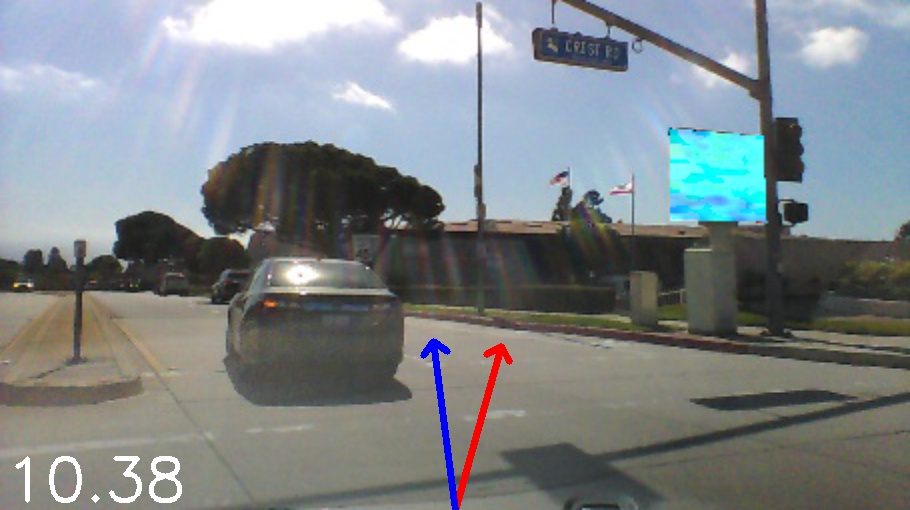} \end{minipage} 
& \begin{minipage}{\figw}
\includegraphics[width=\textwidth,height=\figh]{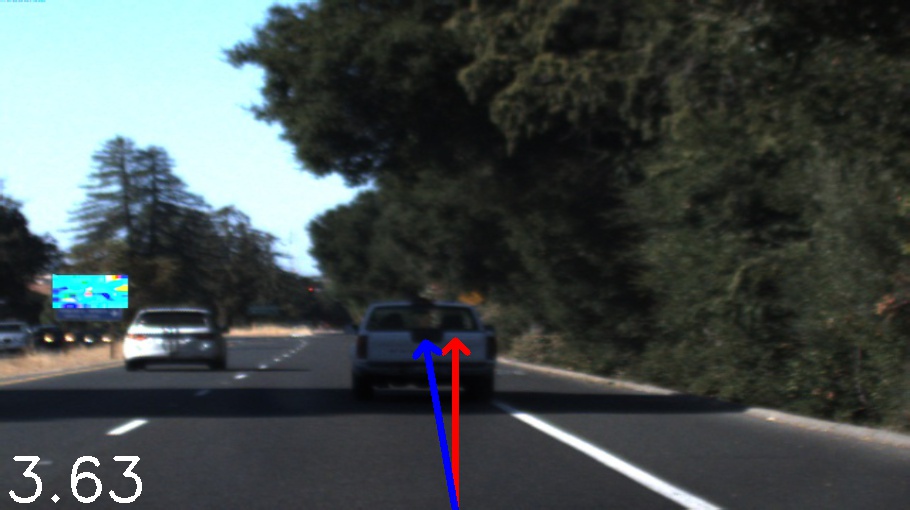} \end{minipage}               
& \begin{minipage}{\figw} 
\includegraphics[width=\textwidth,height=\figh]{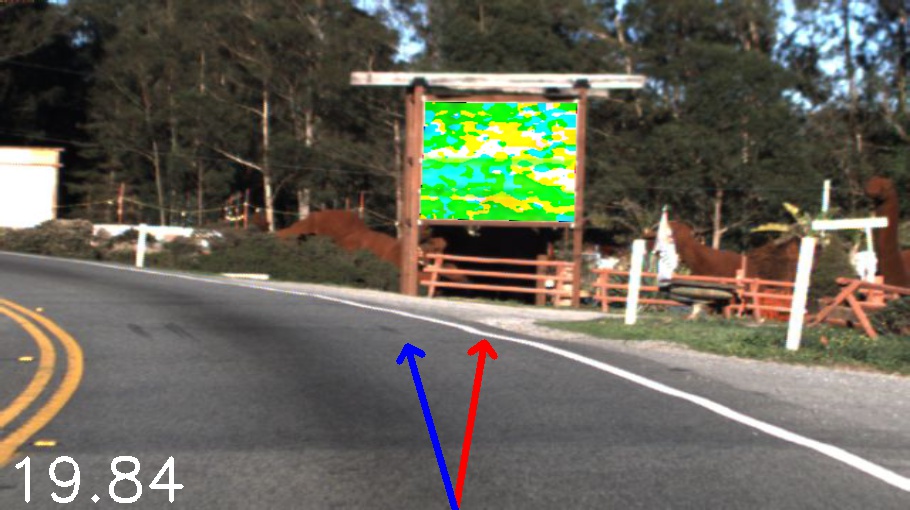} \end{minipage}               
& \begin{minipage}{\figw} \includegraphics[width=\textwidth,height=\figh]{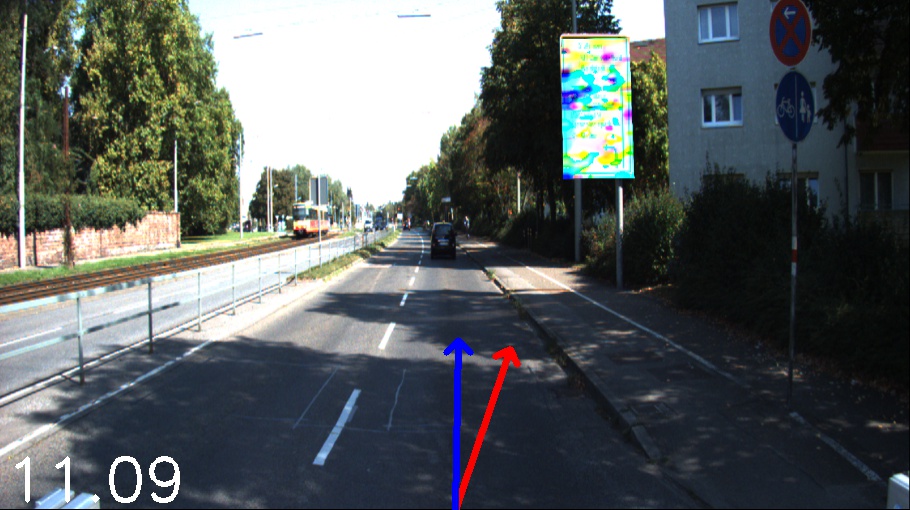} \end{minipage}         
& \begin{minipage}{\figw} \includegraphics[width=\textwidth,height=\figh]{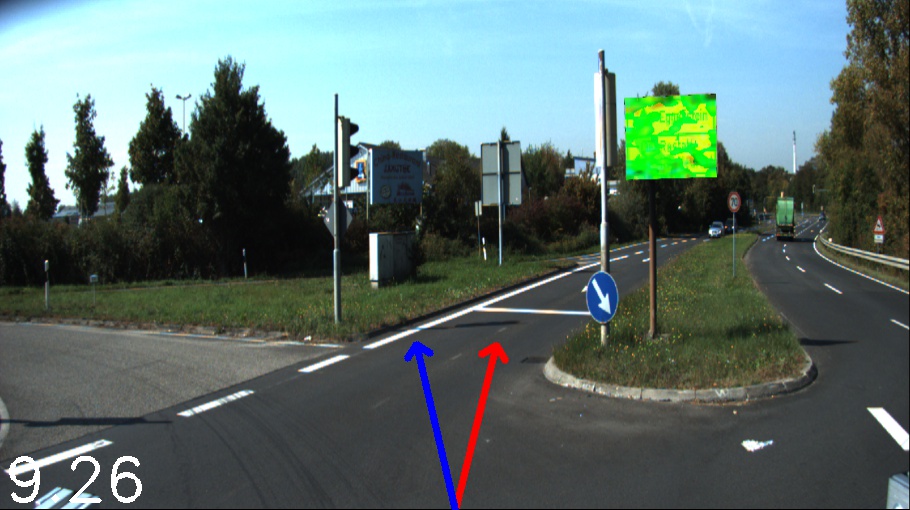} \end{minipage}        
& \begin{minipage}{\figw} 
\includegraphics[width=\textwidth,height=\figh]{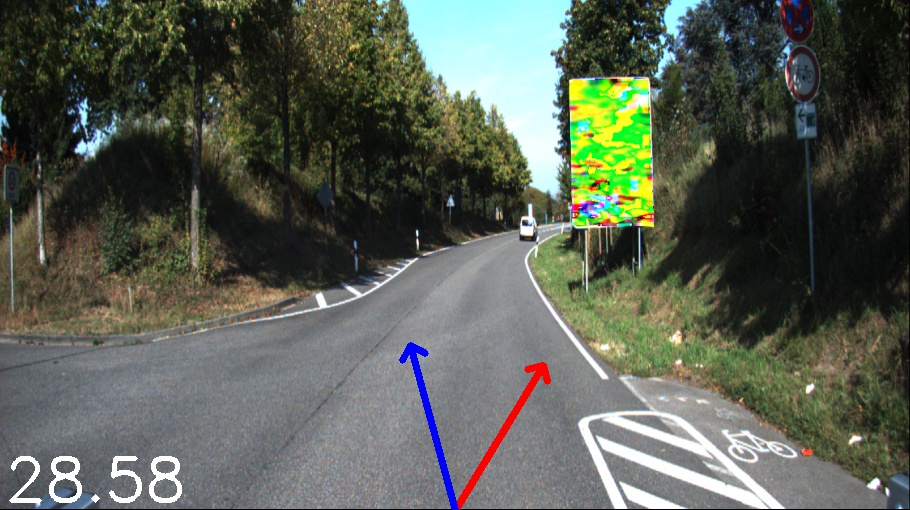} \end{minipage}         
\\ \hline
Dave\_V2 
& \begin{minipage}{\figw} \includegraphics[width=\textwidth,height=\figh]{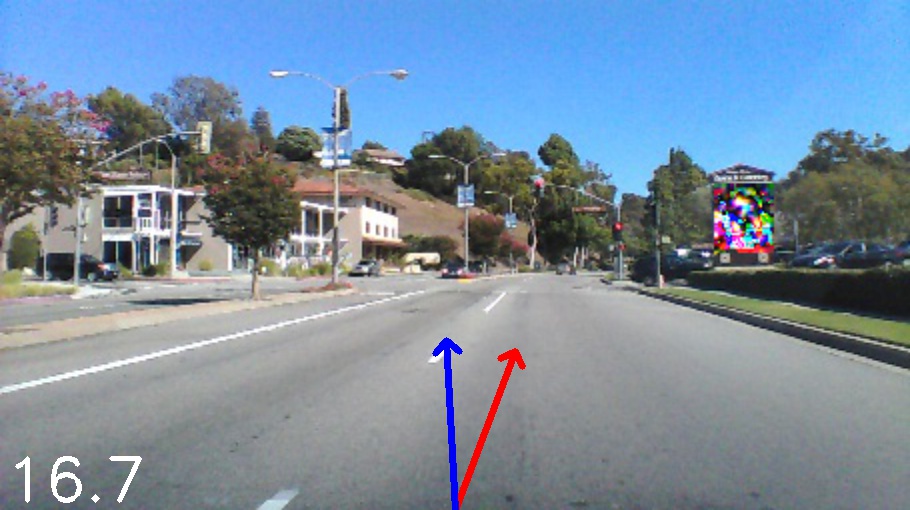} \end{minipage}         
& \begin{minipage}{\figw} 
\includegraphics[width=\textwidth,height=\figh]{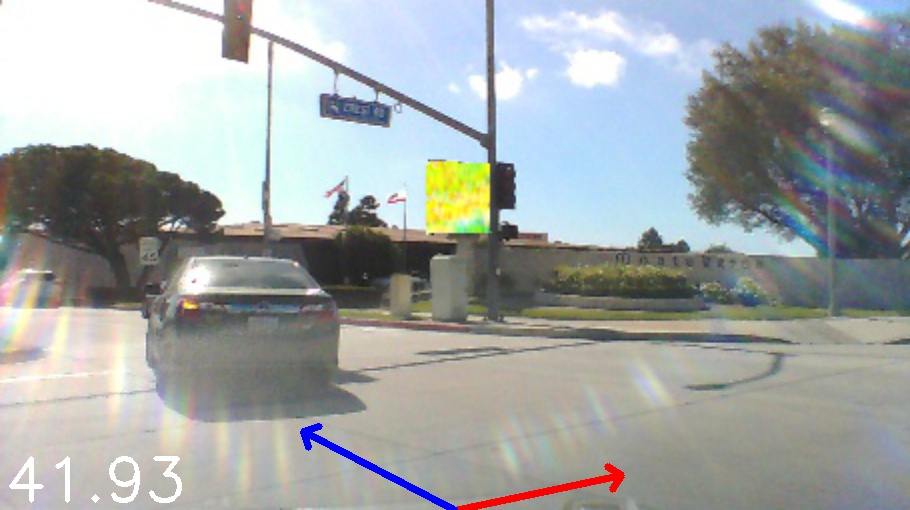} \end{minipage} 
& \begin{minipage}{\figw} \includegraphics[width=\textwidth,height=\figh]{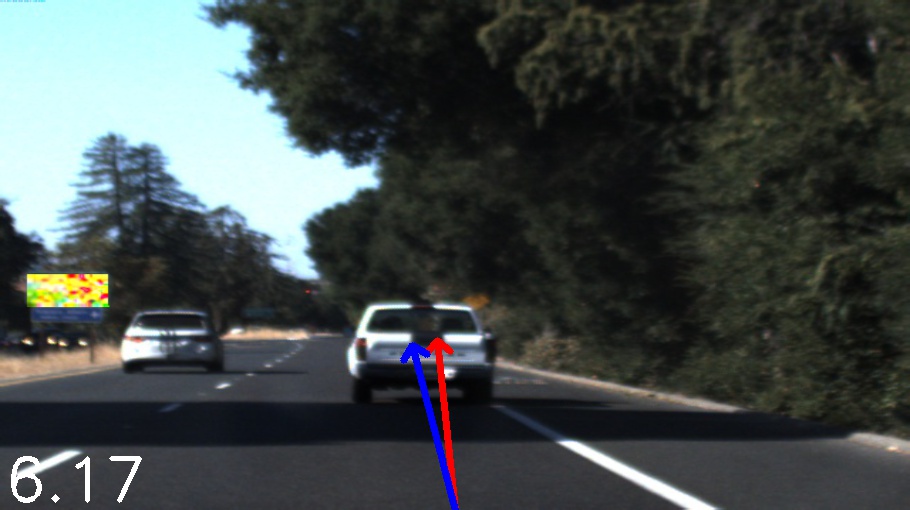} \end{minipage}               
& \begin{minipage}{\figw} 
\includegraphics[width=\textwidth,height=\figh]{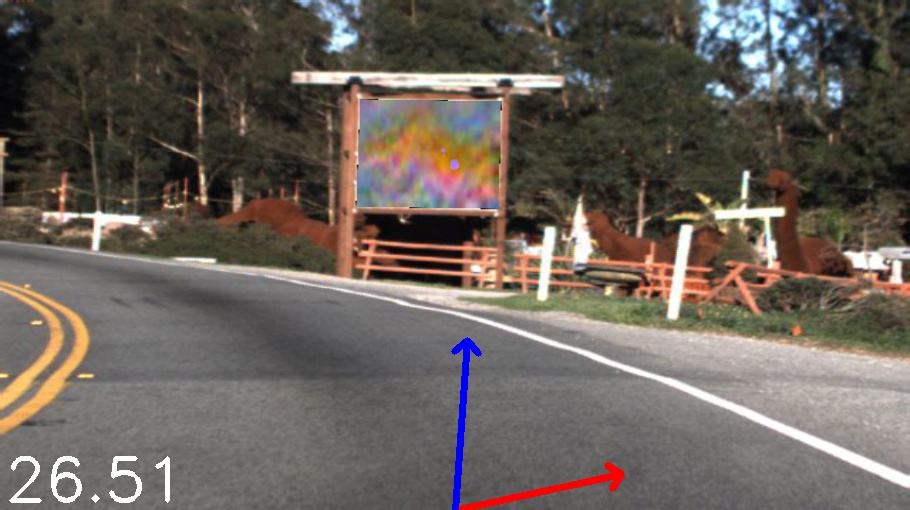} \end{minipage}               
& \begin{minipage}{\figw} \includegraphics[width=\textwidth,height=\figh]{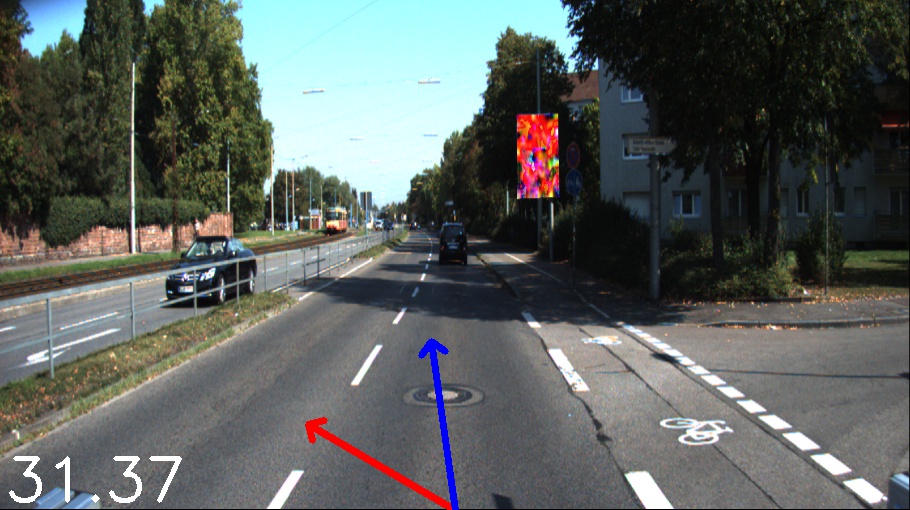} \end{minipage}         
& \begin{minipage}{\figw}
\includegraphics[width=\textwidth,height=\figh]{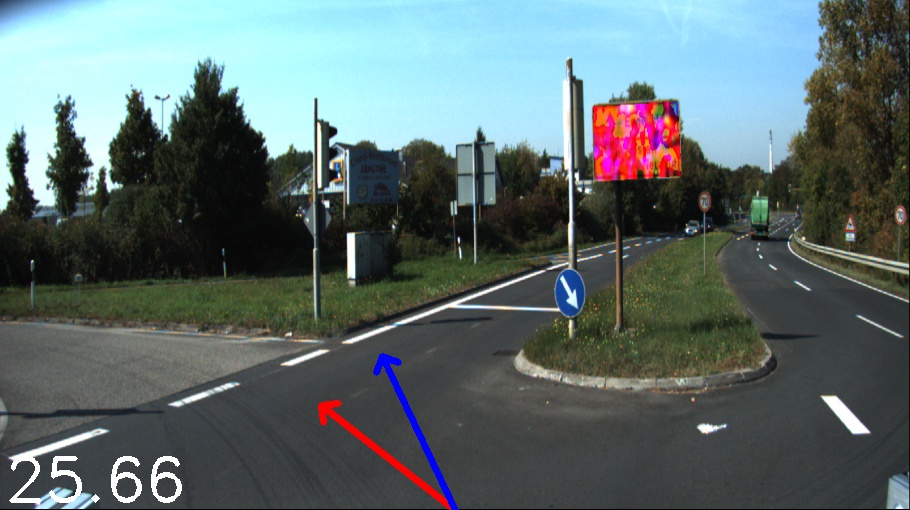} \end{minipage}        
& \begin{minipage}{\figw} 
\includegraphics[width=\textwidth,height=\figh]{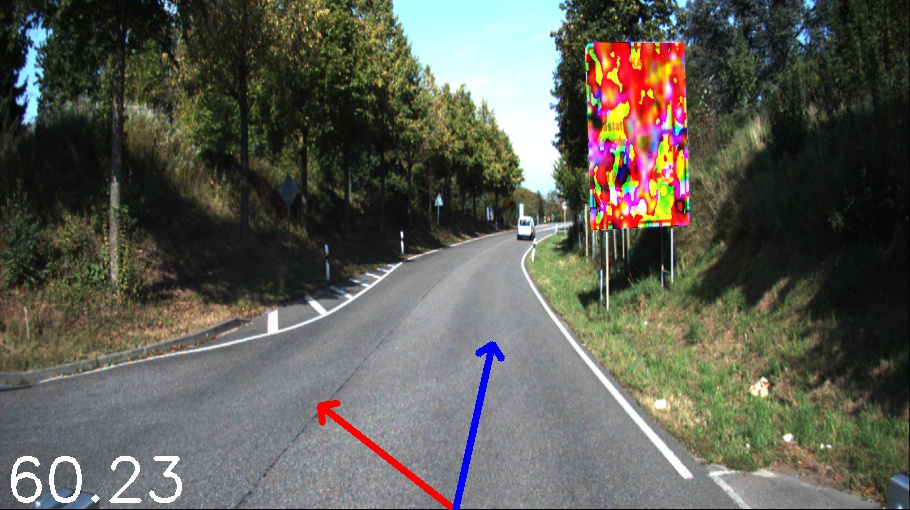} \end{minipage}      
\\ \hline              
Dave\_V3  
& \begin{minipage}{\figw} \includegraphics[width=\textwidth,height=\figh]{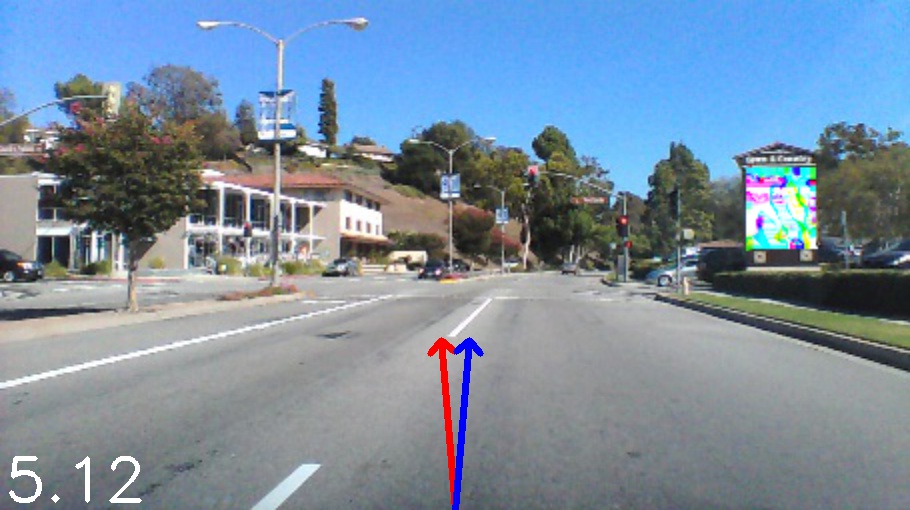} \end{minipage}         
& \begin{minipage}{\figw} 
\includegraphics[width=\textwidth,height=\figh]{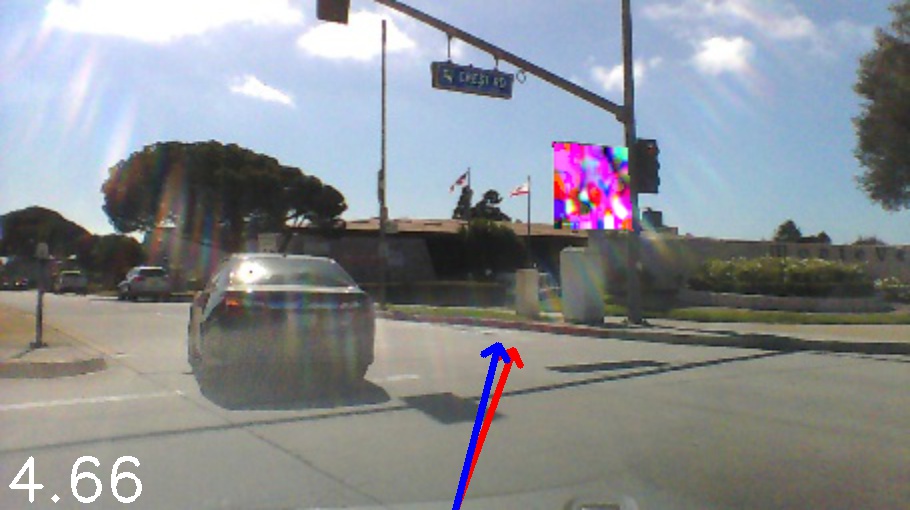} \end{minipage} 
& \begin{minipage}{\figw} \includegraphics[width=\textwidth,height=\figh]{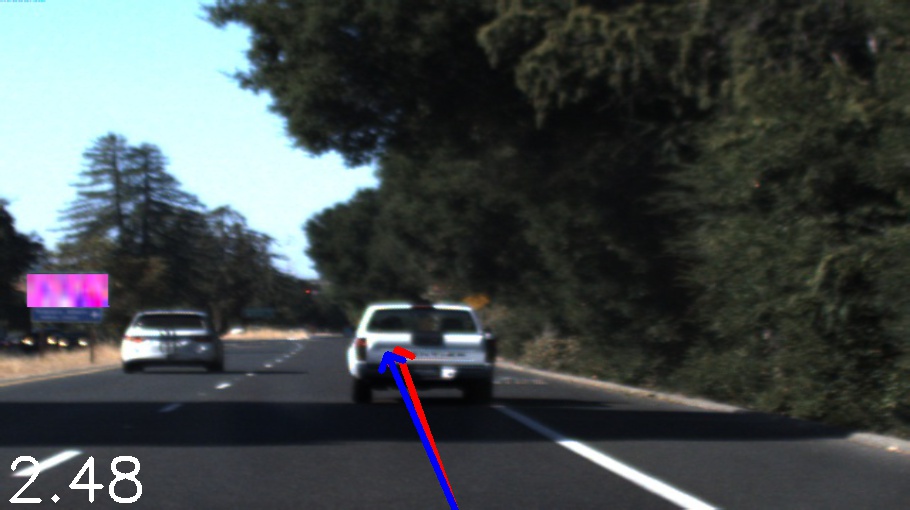} \end{minipage}               
& \begin{minipage}{\figw} 
\includegraphics[width=\textwidth,height=\figh]{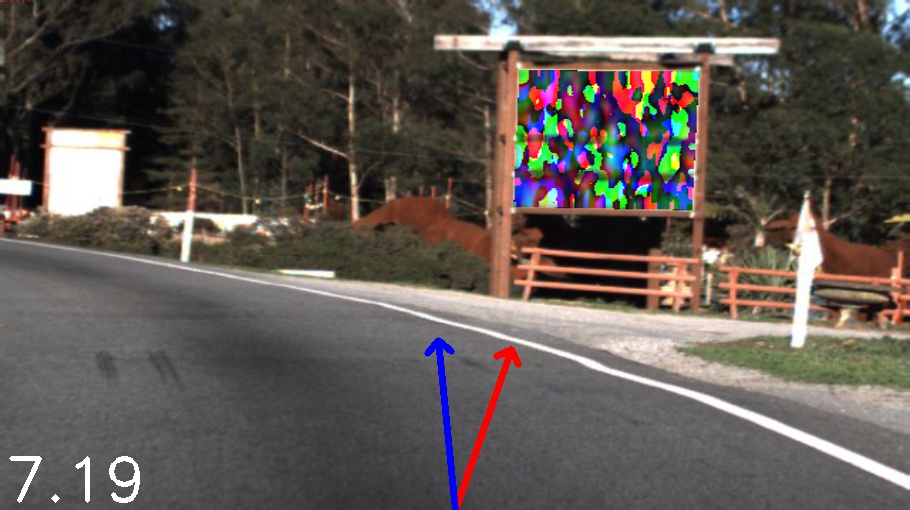} \end{minipage}               
& \begin{minipage}{\figw} \includegraphics[width=\textwidth,height=\figh]{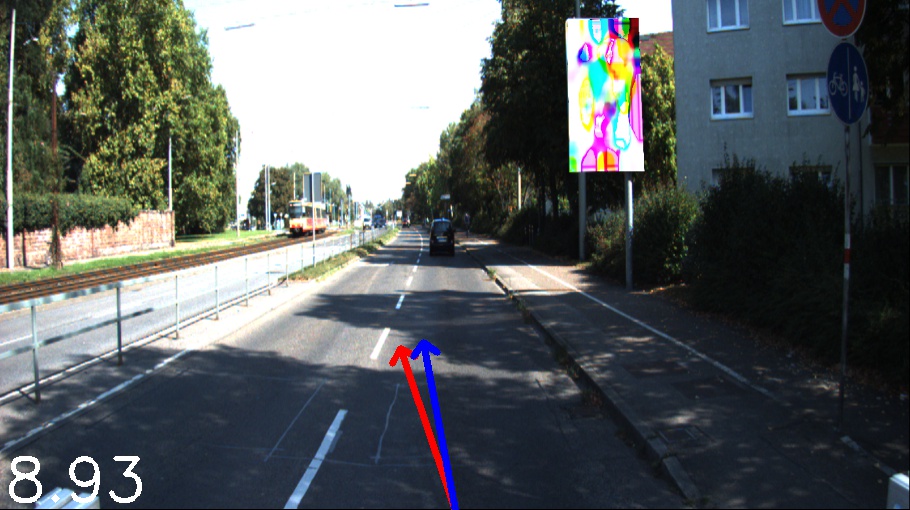} \end{minipage}         
& \begin{minipage}{\figw} \includegraphics[width=\textwidth,height=\figh]{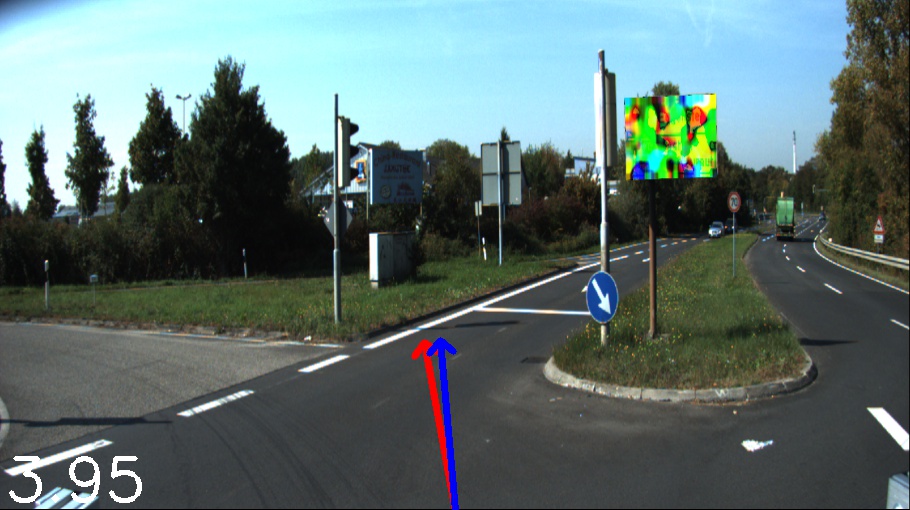} \end{minipage}        
& \begin{minipage}{\figw} 
\includegraphics[width=\textwidth,height=\figh]{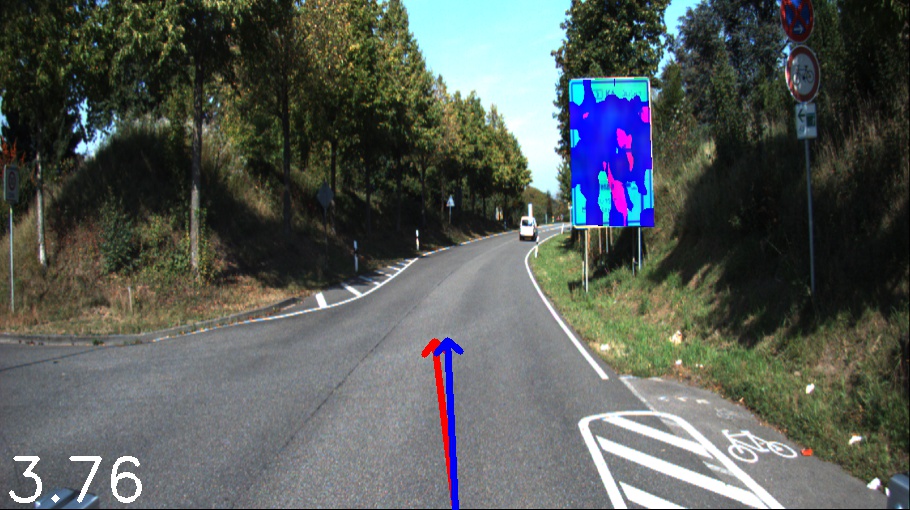} \end{minipage}      
\\ \hline 
Epoch 
& \begin{minipage}{\figw} 
\includegraphics[width=\textwidth,height=\figh]{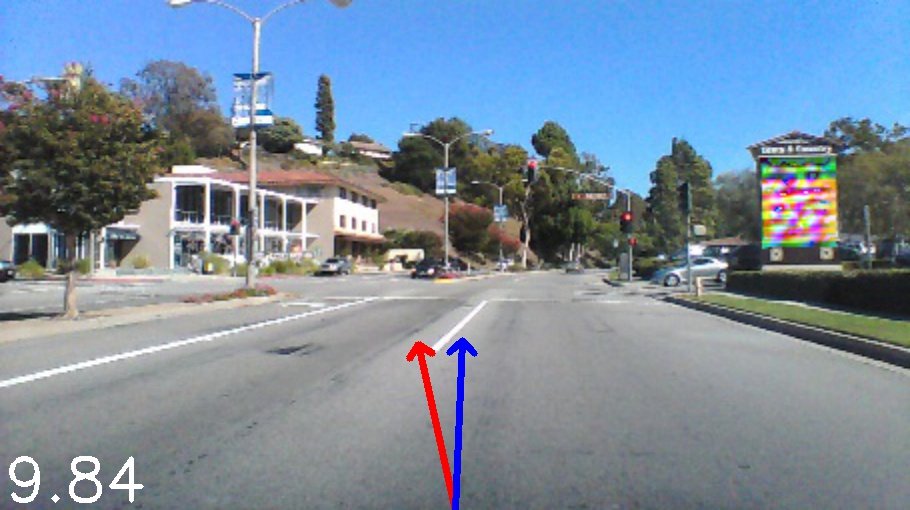} \end{minipage}         
& \begin{minipage}{\figw} 
\includegraphics[width=\textwidth,height=\figh]{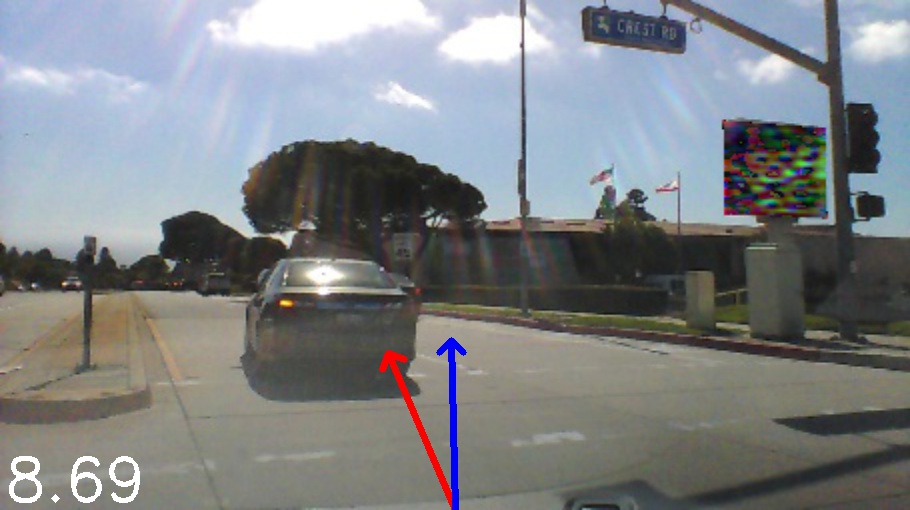} \end{minipage} 
& \begin{minipage}{\figw} \includegraphics[width=\textwidth,height=\figh]{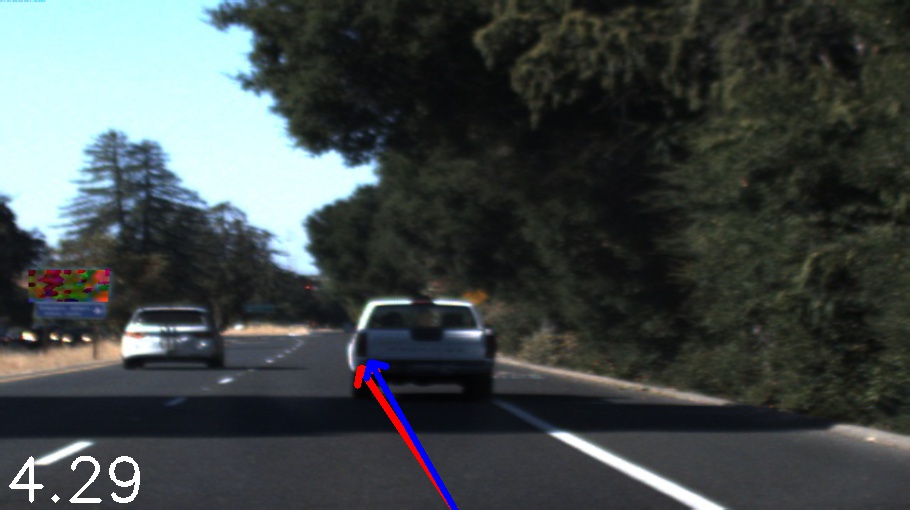} \end{minipage}               
& \begin{minipage}{\figw} 
\includegraphics[width=\textwidth,height=\figh]{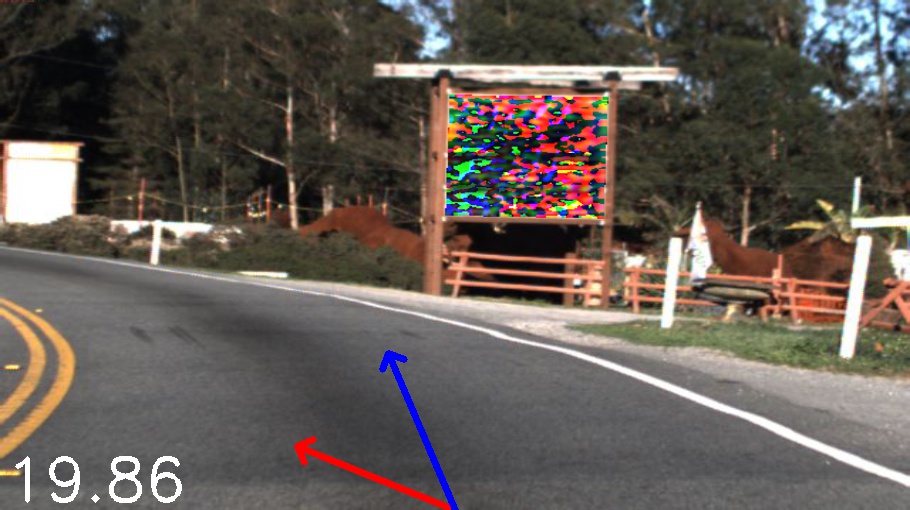} \end{minipage}               
& \begin{minipage}{\figw} \includegraphics[width=\textwidth,height=\figh]{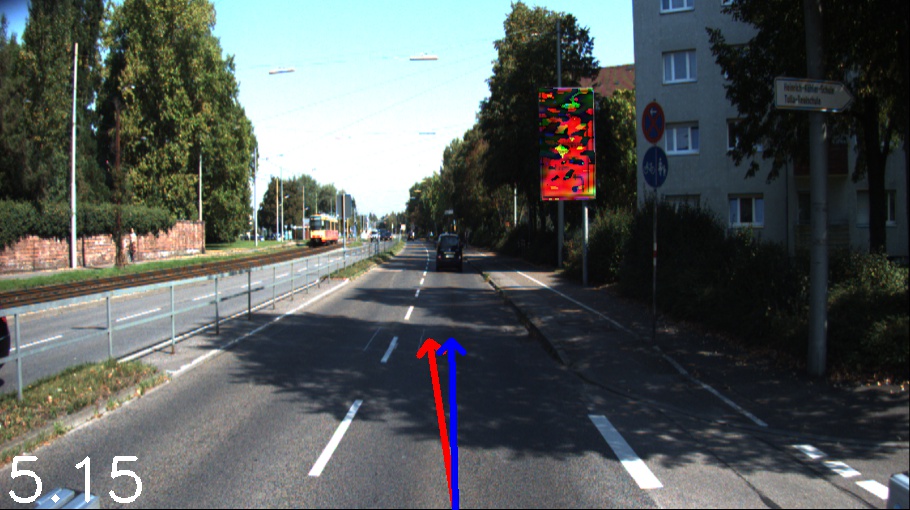} \end{minipage}         
& \begin{minipage}{\figw} \includegraphics[width=\textwidth,height=\figh]{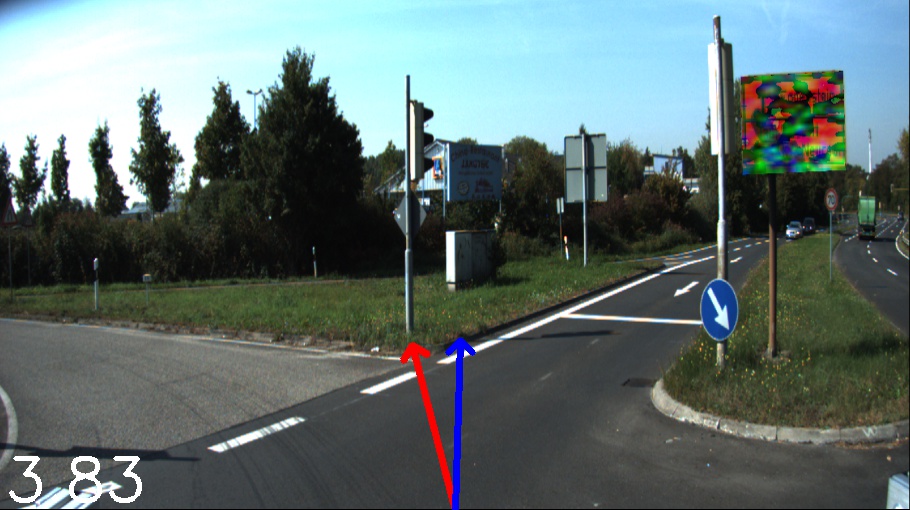} \end{minipage}        
& \begin{minipage}{\figw} 
\includegraphics[width=\textwidth,height=\figh]{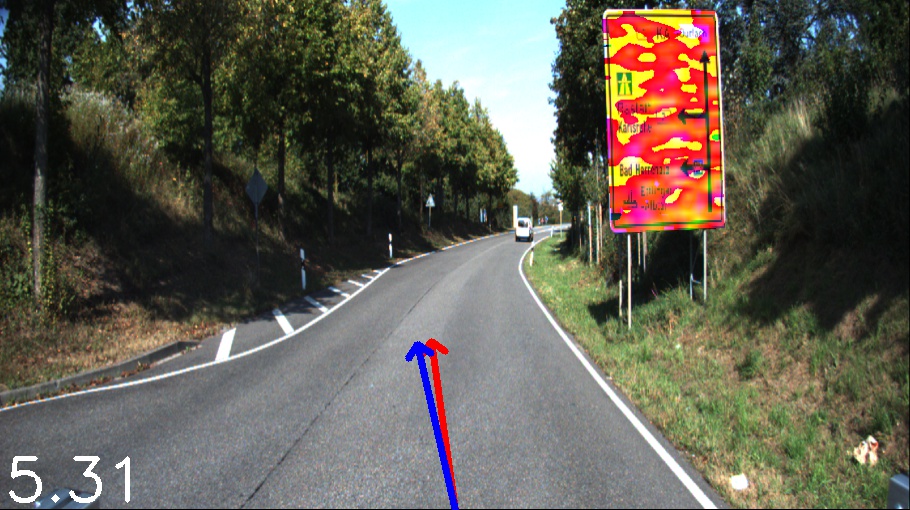} \end{minipage}       
\\ \hline 
\end{tabular}
\end{sidewaystable*}

\vspace{2mm}
\begin{table*}[t]
\centering
    \begin{minipage}[t]{.24\textwidth}
\subfloat[Dave\_Straight1]{\includegraphics[width=1\columnwidth]{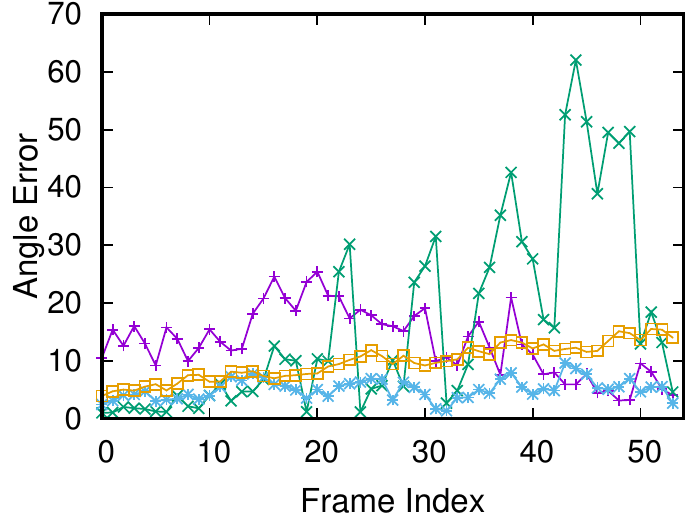}}
    \end{minipage}
        \begin{minipage}[t]{.24\textwidth}
\subfloat[Dave\_Curve1]{\includegraphics[width=1\columnwidth]{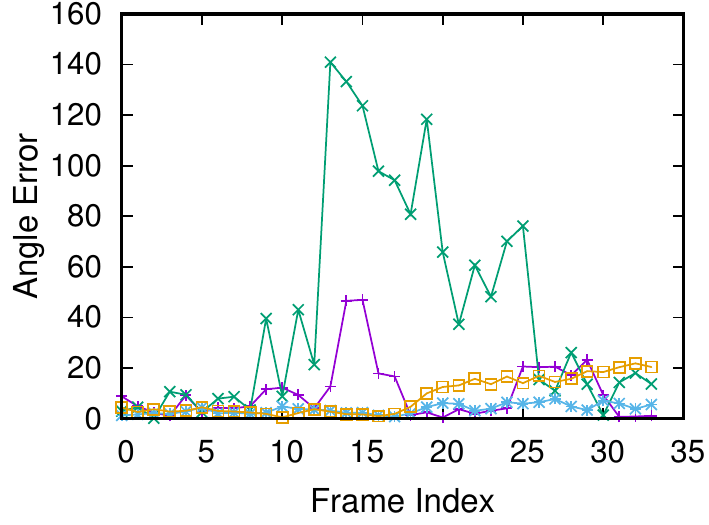}}
    \end{minipage}
    \begin{minipage}[t]{.24\textwidth}
\subfloat[Udacity\_Straight1]{\includegraphics[width=1\columnwidth]{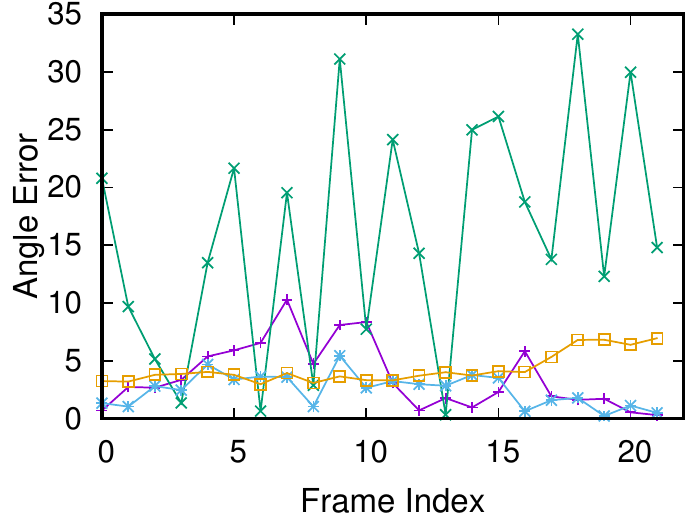}}
    \end{minipage}
        \begin{minipage}[t]{.24\textwidth}
\subfloat[Udacity\_Curve1]{\includegraphics[width=1\columnwidth]{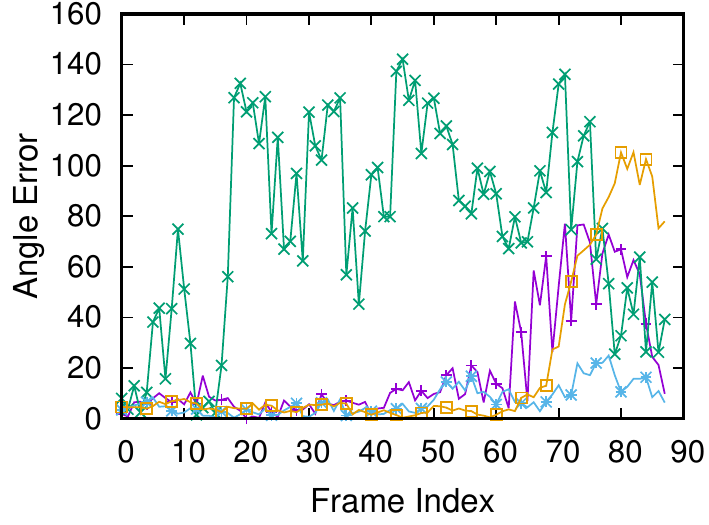}}
    \end{minipage}  
    
        \begin{minipage}[t]{.24\textwidth}
\subfloat[Kitti\_Straight1]{\includegraphics[width=1\columnwidth]{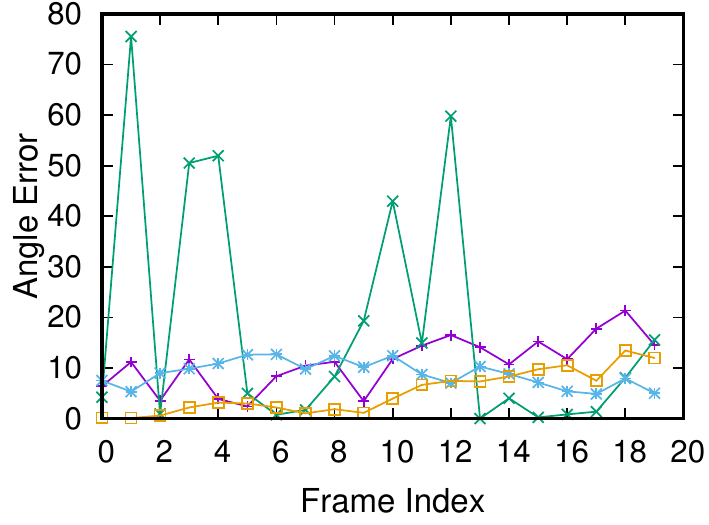}}
    \end{minipage}
        \begin{minipage}[t]{.24\textwidth}
\subfloat[Kitti\_Straight2]{\includegraphics[width=1\columnwidth]{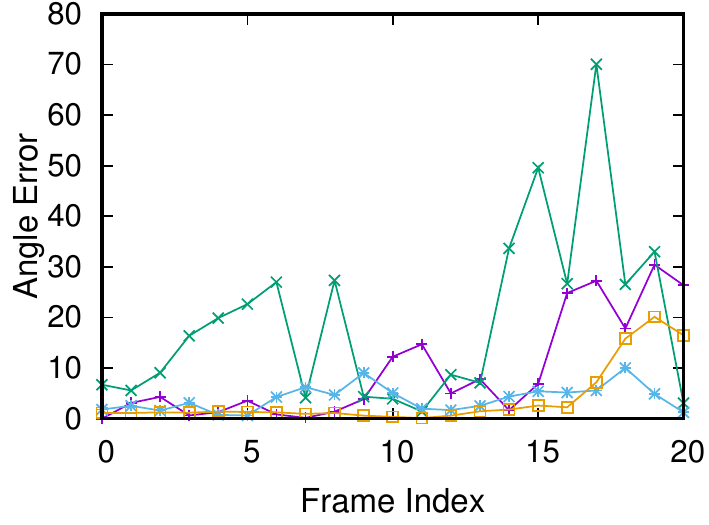}}
    \end{minipage}
    \begin{minipage}[t]{.24\textwidth}
\subfloat[Kitti\_Curve1]{\includegraphics[width=1\columnwidth]{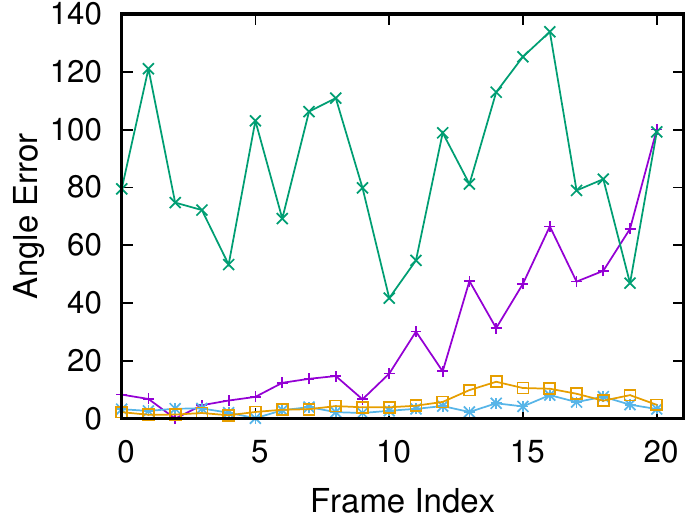}}
    \end{minipage}
 \begin{minipage}[t]{.24\textwidth}
 \centering
 \vspace{8mm}
\subfloat{\includegraphics[width=1in]{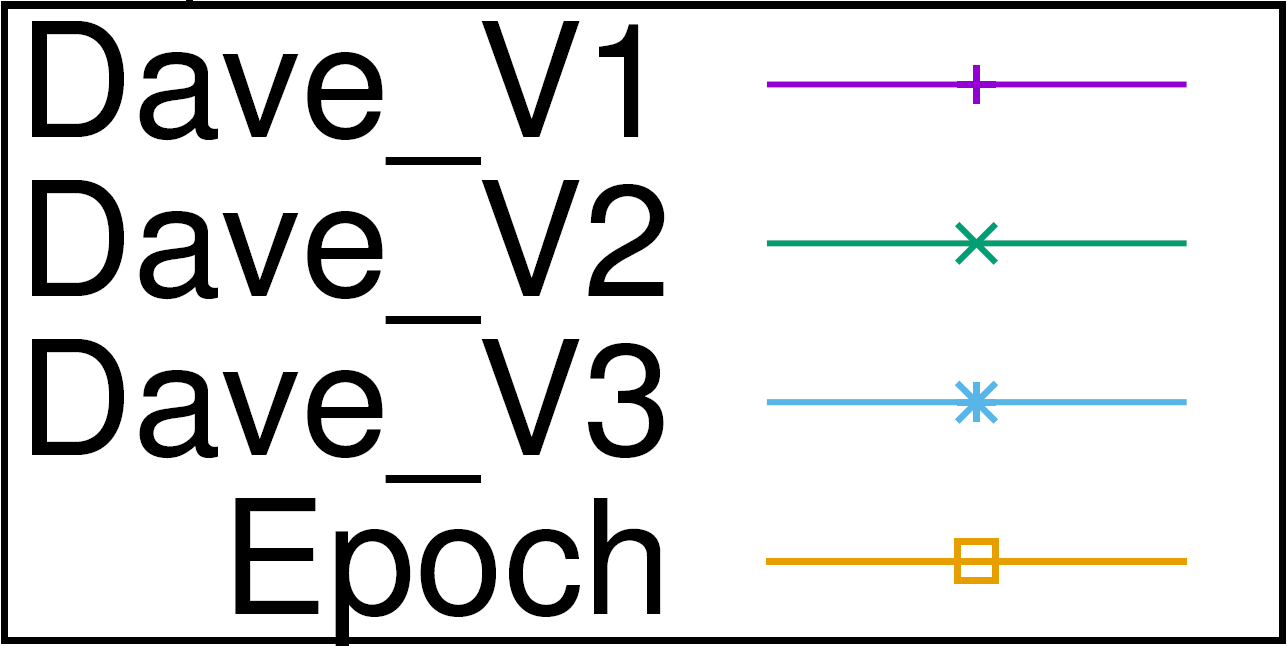}}
    \end{minipage}
\captionof{figure}{Steering angle error variations along the timeline}
\label{fig:timeline}
\end{table*} 


\noindent \textbf{Datasets and Steering Models.}
We use four pre-trained popular CNNs as targeted steering models, which have been widely used in autonomous driving testing~\cite{deeproad,deeptest,deepxplore, deepgauge}. Specifically, we test three models based on the DAVE self-driving car architecture from NVIDIA, denoted as Dave\_V1~\cite{dave1}, Dave\_V2~\cite{dave2}, Dave\_V3~\cite{dave3}, and the Epoch model~\cite{cg32} from the Udacity challenge~\cite{udacity}. 
 Specifically, Dave\_V1 is the original CNN architecture presented in NVIDIA's Dave system~\cite{dave}. Dave\_V2~\cite{dave2} is a variation of Dave\_V1 which normalizes the randomly initialized network weights and removes the first batch normalization layer. 
Dave\_V3~\cite{dave3} is another publicly available steering model which modifies the original Dave model by removing two convolution layers and one fully connected layer, and inserting two dropout layers among the three fully connected layers. 
As the pre-trained Epoch weights are not publicly available, we train it following the instructions provided by the corresponding authors using the Udacity self-driving Challenge dataset~\cite{udacity}.


The datasets used in our experiments include: (1) Udacity self-driving car challenge dataset~\cite{udacity} which contains 101,396 training images captured by a dashboard mounted camera of a driving car and the simultaneous steering wheel angle applied by the human driver for each image; (2) Dave testing dataset~\cite{dave-data} which contains 45,568 images recorded by a GitHub user to test the NVIDIA Dave model; and (3) Kitti~\cite{kitti} dataset which contains 14,999 images from six different scenes captured by a VW Passat station wagon equipped with four video cameras.

The dataset used for our physical case studies consists of videos recorded by a tachograph mounted behind the windshield of a driving car for driving by a pre-placed roadside billboard on campus. We use aforementioned pre-trained steering models to predict every frame, and use the resultant steering angle decisions as the ground truth.

\vspace{1mm}
\noindent \textbf{Experiment Design.} Based on our discussion from Section~\ref{sec:metrics}, we evaluate the efficacy of our algorithm by measuring the Average Angle Errors of all frames in a scene, both digitally and physically. 

For digital tests, our scene selection criteria is that the billboard should appear entirely in the first frame with more than 400 pixels and partially disappear in the last frame. We then randomly select seven scenes that satisfy this criteria from aforementioned datasets, and evaluate on all the selected scenes. The selected scenes in each dataset cover both straight and curved lane scenarios. Since all these datasets do not contain coordinates of billboards, we have to label the four corners of billboards in every frame of the selected scenes. To make the labeling process semi-automated, we use the motion tracker functionality of Adobe After Effects~\cite{aftereffects} to automatically track the movement of billboard's four corners among consecutive frames. We then perform necessary adjustments for certain frames whose coordinates are not accurate enough.
We list the statistics about all the studied scenes in Table~\ref{tb:scenes}, where the first column lists the names of scenes, the second column shows the number of images in every scene, the third to fifth columns indicate the resolutions of images and the min/max sizes of billboards in each scene. 
In digital tests, there is no color adjustment under different environmental conditions. The final adversarial example is patched into every frame according to the projection function. Then we use the steering models to predict the patched images and compare them against the ground-truth steering decisions recorded in the given datasets.

Our compared baseline is the inference steering angle for each given trained model. Our approach seeks to maximize the distance from the baseline, regardless whether baseline is ground truth or inference results. We choose to present results associated with using inference as the baseline because the driving datasets used in experiments may not have ground truth steering angle. Since there exists many physically-correct driving behaviors, we use statistical methods (i.e., average, percentage) to test the effectiveness during the entire driving segment rather than within each individual frame.

For physical tests, we record multiple videos using a tachograph mounted on a vehicle at various realistic driving speeds. We place a billboard alongside the road, and drive towards the billboard straightly along the central of the road. We start recording at approximately 100 ft away from the billboard, and stop recording once the vehicle passes the billboard. We perform multiple physical tests under three different weather conditions including sunny, cloudy, and dusk weather. The physical test is composed of the following two phases: 
\begin{itemize}
\item(1) Phase I: We use a white billboard with its four corners painted as black and then use a golden billboard with four blue corners. For each board, we record and drive along the central of the road with a slow speed of 10mph in order to capture sufficient frames (i.e., training videos).

\item(2) Phase II: We send the input videos to our testing algorithm to automatically generate the adversarial perturbation, which will then be pasted on the billboard. We then drive by the adversarial billboard with normal speed at 20 mph and record the video (i.e., testing video). We calculate the average angle error compared to the ground truth steering angle for every frame of the video.
\end{itemize}

We note that strictly speaking, a real-world test would involve actually autonomous vehicles driving by the billboard to observe the adversarial impact. Unfortunately, due to lacking actualy autonomous vehicles, to validate \tech in real-world settings, we took a similar approach applied in the following state-of-the-art autonomous-driving  research~\cite{deepxplore,deeptest,deeproad}, which also has not involved actual autonomous vehicles in the evaluation. Specifically, we take videos with different driving patterns as inputs, which can be as exhaustive as possible to cover all potential viewing angles at different vehicle-to-billboard distances. In the physical-world evaluation, we tried to pre-record as many abnormally-driving videos as possible to cover a majority of the possible misled driving scenarios of an actual autonomous vehicle. Such videos have been applied in the adversarial construction/training phase. We show such an abnormally-driving video in the following link: https://github.com/evilbillboard/EvilBillboard. This video shows that \tech is able to continuously deviate a car within each frame. This would mimic one of the many actual autonomous driving scenarios where the vehicle is continuously misled by \tech at each frame (i.e., the misled angle within each frame is similar to the one shown in this video).

\subsection{Digital Perturbation Results}

The results of digital perturbations are shown in Table~\ref{tb:overall}, where each column represents a specific scene, and each row represents a specific steering model.
Every image in a cell shows a representative frame that has the \emph{median} steering angle divergence. For example, the image in cell (Dave\_V1, Udacity\_Scene1) represent the image in Udacity dataset Scene1 has the Average Angle Error among all frames in the same scene when predicted by Dave\_V1 steering model.
Two arrows shows the steering angle decision divergence in each image, where the blue one is the ground truth and the red one is the steering angle of the generated adversarial examples. 
We observe that in all scenes, \tech~makes all steering models generate observable average steering angle divergences. Specifically, \tech~misleads the Dave\_V1 model by more than $10^{\circ}$ in 6 out of 7 scenes, except for Kitti\_Straight1 in which the billboard occupies a small space. Dave\_V2 incurs the largest average divergence -- more than $16.7^{\circ}$ among all scenes.\remove{The reason is that Dave\_V2 model is not robust enough, causing its baseline prediction without perturbation to have non-negligible divergence compared to the ground truth.} 
The test cases of Dave\_V2 model show that even with underfitted model, \tech~can still greedily enlarge such divergence. \tech~causes the smallest divergence for the Dave\_V3 model -- $0.44^{\circ} - 25.01^{\circ}$. 
The reason is because Dave\_V3 introduces three dropout layers between four fully connected layer, and use augmented training data, which both contribute to the enhanced robustness and generalization of the trained model. Particularly, the adoption of dropout layer which randomly deactivates half of the neurons, can cause part of the perturbations on billboards being deactivated, thus reducing the efficacy of adversarial perturbations. We note that the Epoch model also adopts dropout layers, so its average angle error is also small compared to Dave\_V1 and Dave\_V2 in all scenes. 
However, Epoch does not apply the training data augmentation used by Dave\_V3 which crops the images to train only the road pavement, thus the perturbations on the roadside billboard has more influence to the prediction compared to Dave\_V3, resulting in a larger average angle error.

We further show the results on steering angle error along the timeline for each studied scene, from the first frame to the last frame where the billboard size increases monotonically among these frames. The results are shown in Fig.~\ref{fig:timeline}, where each sub-figure indicates a specific scene, the x-axis is the indexes of images along the timeline, and the y-axis is the steering angle error ($\circ$). We observe that in most scenes the steering angle errors increase when the billboard size increases, as indicated by the Dave\_V1 lines shown in Fig.~\ref{fig:timeline} (d), (e), (f), (g). The reason behind is intuitive -- larger billboards in images may activate stronger perturbations. 
On the contrary, certain lines do not follow this trend, as indicated by Fig.~\ref{fig:timeline} (a), (b), (c). For example, in Fig.~\ref{fig:timeline} (b), frames in the middle contribute more steering angle errors for the Dave\_V1 model. 
We learn that in such scenarios, even though the billboard is quite small in the image, it can still lead to large steering angle divergence when applying adversarial perturbations, indicating the test effectiveness and robustness of \tech. 


\subsection{Parameter Tuning}

\begin{figure}[t]
\centering
\subfloat[]{\includegraphics[width=.5\columnwidth]{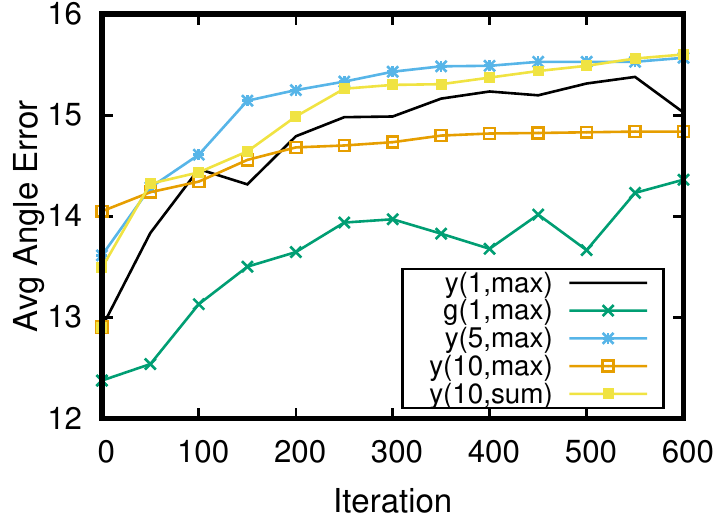}}
\subfloat[]{\includegraphics[width=.5\columnwidth]{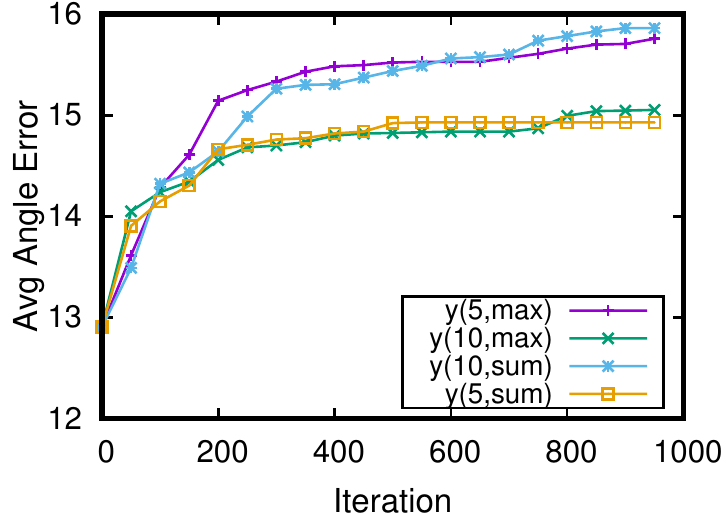}}
\caption{Convergence of AAE w.r.t different parameters.}
\label{fig:tune}
\end{figure}

In this set of experiments, we show that how parameter tuning may affect the AAE--average angle error.
Fig.~\ref{fig:tune} shows the convergence trend when applying different parameters. The x-axis is the enhanced iteration, y-axis is AAE, and the lines represent different parameter settings. For example, line $y$ indicates that the iterations begin with initializing the billboard as yellow, $y(5)$ indicates setting the batch size as 5. Similar settings apply to $y(10)$, and $g$ indicates an initialized green billboard. Line $y(10, sum)$ indicates that besides using batch size 10, it also uses sum to update gradient, instead of the default max pooling. We observe from Fig.~\ref{fig:tune}(a) that starting iterations from yellow is overall better than starting from green in this example. Additionally, we observe that two lines behave much better than other lines -- y(5, max) and y(10, sum). 
To further explore the tradeoff of batch size and sum/max method, we conduct another set of experiments which iterate up to 1000 iterations, whose results are shown in Fig.~\ref{fig:tune}(b).
We observe that, two lines representing $y(5,max)$ and $y(10,sum)$ outperform the other two lines. What we learn from these two figures are: (1) carefully choosing the initial color of the billboard can efficiently increase the converge speed and yield a better results; and (2) there is no clear indications showing there exists a better parameter choice between choosing a large or small batch, and choosing max or sum to update gradient.

\begin{figure}[t]
\centering
\subfloat[]{\includegraphics[width=.5\columnwidth]{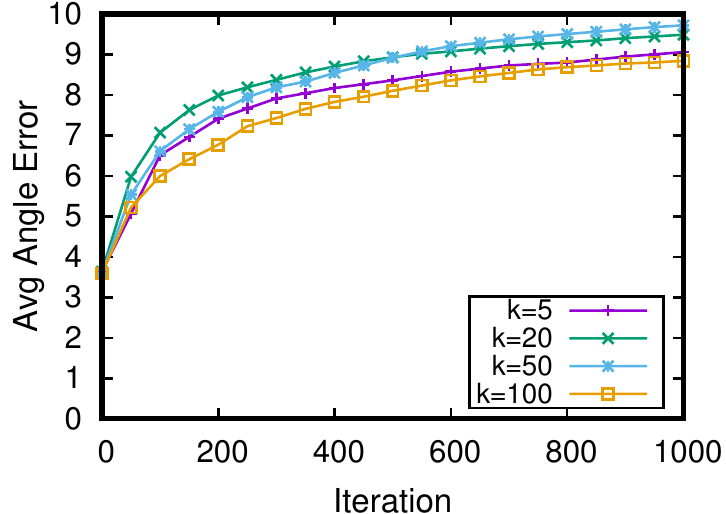}}
\subfloat[]{\includegraphics[width=.5\columnwidth]{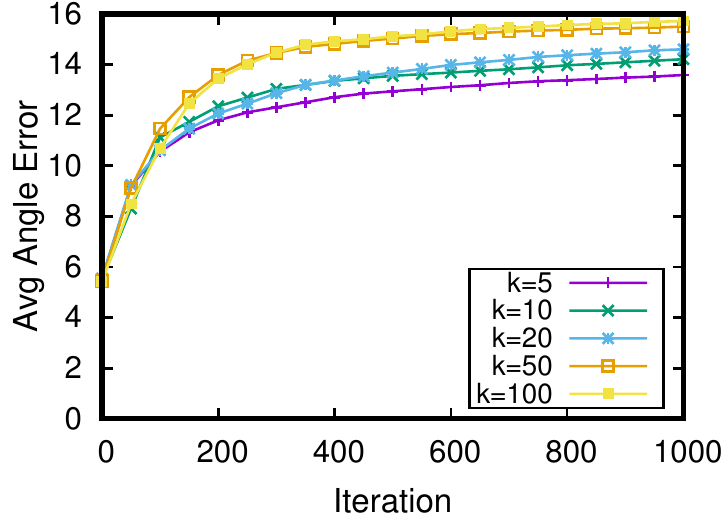}}\\
\subfloat[]{\includegraphics[width=.5\columnwidth]{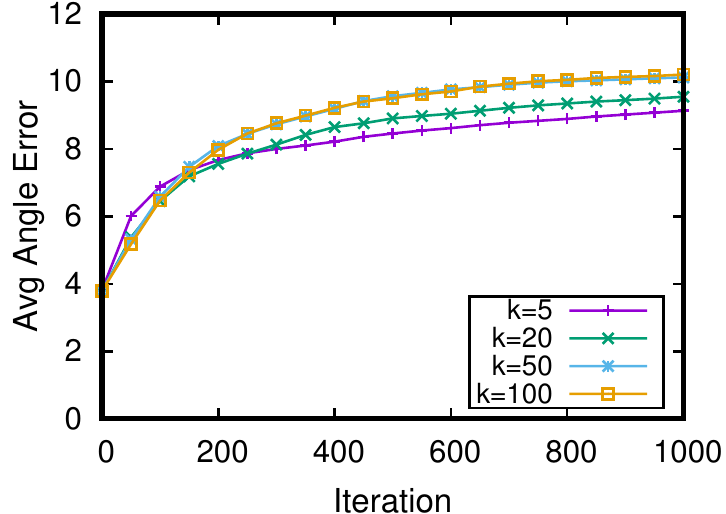}}
\subfloat[]{\includegraphics[width=.5\columnwidth]{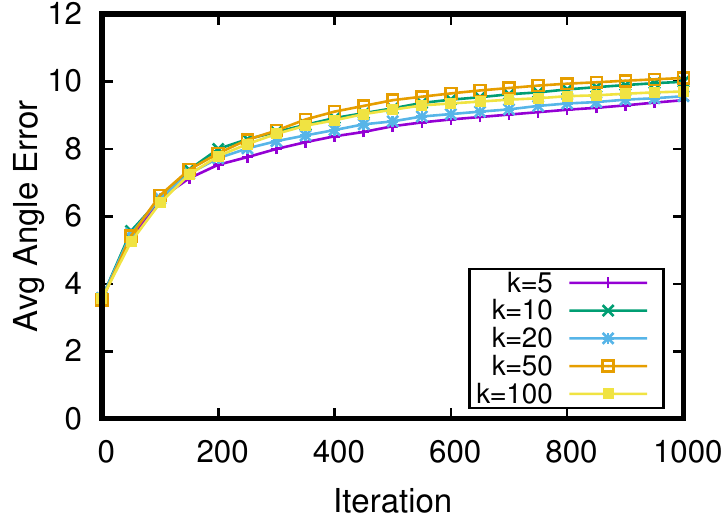}}
\caption{Converge over iterations with different parameter tunings: (a) first 40 frames; (b) 40 frames in the middle; (c) interleaving half frames; (d) all frames.}
\label{fig:pov}
\end{figure}

To figure out how the training set affects the convergence and the objective, we use the same initial color, batch size and overlapping handling ($y(5,max)$) for different subsets among the total 80 frames in Udacity\_Curve1. The results are shown in Fig.~\ref{fig:pov}, where four sub-figures represent (a) the first 40 frames, (b) the last 40 frames, (c) the 40 frames with even indexes, and (d) all 80 frames, respectively. Lines in each sub-figure represent different $k$ values to be updated. We observe that all lines ascend fast at early iterations, and the increase rates drop after around 400 iterations. The lines in Fig.~\ref{fig:pov} (a) converge to a lower AAE compared to lines in the other three sub-figures. Lines using the last 40 frames clearly achieve better results. From this observation, we learn that the chosen training set does affect the final objective in the sense that images with larger billboards can achieve better results. Additionally, a larger $k$ value usually achieves better results and faster convergence in most scenarios except for Fig.~\ref{fig:pov} (a). The reason is that in this specific scenario, the billboard occupies a rather small number of pixels. Thus, aggressively increasing the number of updated pixels would cause severe interferences among frames, thus leading to lower AAE. From the parameter tuning experiments, we learn that choosing images with larger billboard space, aggressively updating more pixels, would result in faster convergence and better results. 

\newcommand{\pfigw}{.16\textwidth}
\newcommand{\pfigh}{3cm}
\newcolumntype{P}[1]{>{\hspace{0pt}}p{#1}}

\begin{sidewaystable*}
\caption{Illustration of physical billboard perturbation. (The table is rotated clockwise 90 degree specifically for  better presentation clarity.)}
\label{tb:physical}
\centering
\begin{tabular}{|P{1.7cm}|c|cccc|}\hline
Distances & Perturbation & 100' & 60' & 20' & 10' \\ \hline \hline
White & N/A & 
\begin{minipage}{\pfigw}
      \includegraphics[width=\linewidth,height=\pfigh]{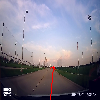}
    \end{minipage} & 
\begin{minipage}{\pfigw}
      \includegraphics[width=\linewidth,height=\pfigh]{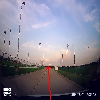}
    \end{minipage} & 
\begin{minipage}{\pfigw}
      \includegraphics[width=\linewidth,height=\pfigh]{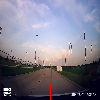}
    \end{minipage} & 
\begin{minipage}{\pfigw}
      \includegraphics[width=\linewidth,height=\pfigh]{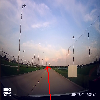}
    \end{minipage}  \\ \hline
Adversarial (left) & \begin{minipage}{\pfigw}
      \includegraphics[width=\linewidth,height=\pfigh]{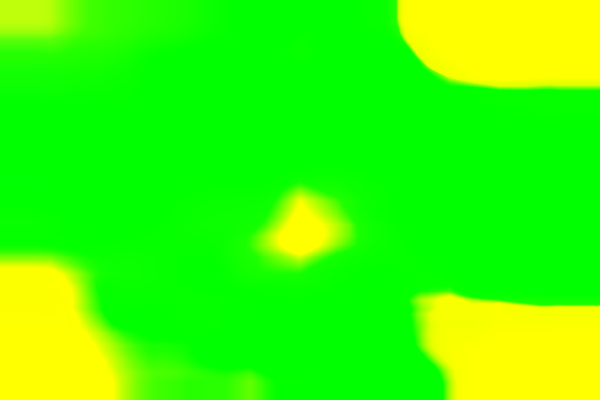}
    \end{minipage} & 
\begin{minipage}{\pfigw}
      \includegraphics[width=\linewidth,height=\pfigh]{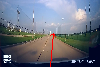}
    \end{minipage} & 
\begin{minipage}{\pfigw}
      \includegraphics[width=\linewidth,height=\pfigh]{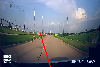}
    \end{minipage} & 
\begin{minipage}{\pfigw}
      \includegraphics[width=\linewidth,height=\pfigh]{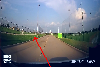}
    \end{minipage} & 
\begin{minipage}{\pfigw}
      \includegraphics[width=\linewidth,height=\pfigh]{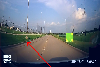}
    \end{minipage} 
 \\ \hline
Adversarial (right) & \begin{minipage}{\pfigw}
      \includegraphics[width=\linewidth,height=\pfigh]{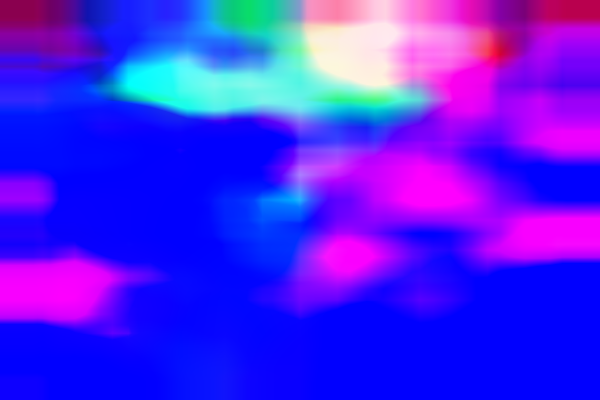}
    \end{minipage} & 
\begin{minipage}{\pfigw}
      \includegraphics[width=\linewidth,height=\pfigh]{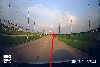}
    \end{minipage} & 
\begin{minipage}{\pfigw}
      \includegraphics[width=\linewidth,height=\pfigh]{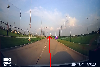}
    \end{minipage} & 
\begin{minipage}{\pfigw}
      \includegraphics[width=\linewidth,height=\pfigh]{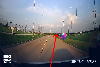}
    \end{minipage} & 
\begin{minipage}{\pfigw}
      \includegraphics[width=\linewidth,height=\pfigh]{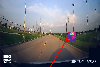}
    \end{minipage}
 \\ \hline
\end{tabular} 
\end{sidewaystable*}

\subsection{Physical Case Study}

As described in Section~\ref{sec:setup}, our physical case study is composed of two phases.
Specifically, for both training and testing videos, we start recording at 100 ft far away and stop recording when the vehicle physically passes the billboard. The driving speed is set to be 10mph for training videos in order to capture sufficient images, and the speed for the testing video is 20mph to reflect ordinary on-campus driving. 
We perform our physical tests on a straight lane without curves under three different weather conditions including sunny, cloudy, and dusk weather. To make the training robust, we record three training videos through three slightly different routes: central, left-shifting, and right-shifting. The billboard used in our experiment has a size of $6' \times 4'$. We adopt $Dave\_V1$ as the steering model.

\begin{table}[]
\vspace{4mm}
\caption{Test effectiveness of physical case study}
\label{tb:physical_values}
\centering
\begin{tabular}{|l|c|c|c|c|c|c|}
\hline
\multicolumn{1}{|c|}{\multirow{2}{*}{Cond.}} & \multicolumn{3}{c|}{Ad\_left} & \multicolumn{3}{c|}{Ad\_right} \\ \cline{2-7} 
\multicolumn{1}{|c|}{}                       & Exp      & Test      & M1     & Exp      & Test      & M1      \\ \hline
Sunny                                        &   23.34       &    8.88       &  32\%  &       -20.09   &      -15.49     &   19\%      \\ \hline
Dusk                   &    42.27      &     26.44      &     100\%   &       -26.54   &      -8.45     &    0     \\ \hline
Cloudy                                       &    9.66      &    4.86       &    0  &        -20.4  &       -11.37    &     26\%  \\ \hline
\end{tabular}
\vspace{4mm}
\end{table}

We define \textit{Exp\_AAE} to indicate the expected average angle error according to the training videos, which is the $M_0$ metric defined in Section~\ref{sec:metrics} based on \emph{digital perturbations}. We use \textit{Test\_AAE} to indicate the actual average angle error for all images in the testing video, which is the $M_0$ metric defined in Section~\ref{sec:metrics} for \emph{physical perturbations}.  
We also record the $M_1$ metric defined in Section~\ref{sec:metrics} for the test video.
We set the steering angle error threshold to $19.8^{\circ}$ since when the driving speed is 20mph, such mis-steering would cause at least an off-track distance of one meter within a time interval of 0.33 second (duration of 10 frames for a 30 FPS camera), which is large enough for causing dangerous driving behaviors as demonstrated by NVIDIA Dave~\cite{dave}.

We note that our chosen evaluation metrics using average angle error and percentage of large angle error can reasonably reflect the overall possibility and strength of misleading for consecutive frames. In the physical experiment, we calculate the angle error threshold according to the speed, which can cause at least one-meter off-tracking (defined as dangerous driving behaviors by NVIDIA Dave2). 

The visible results are shown in Table~\ref{tb:physical}, where each row shows a sunny scene of a testing video, including one video with empty billboard and two videos with adversarial billboards. The second column shows the printable perturbations. Columns 3-6 present different distances between the vehicle and the billboard. We observe that, with white billboard, the steering angles are almost straight in all distances. With the first (bright) adversarial billboard, the steering angles turn left to a certain degree; on the contrary, the second (dark) adversarial billboard leads steering to the right. As mentioned in Sec.~\ref{sec:design}, this is controlled by setting gradient flag(+/-).

The values of test effectiveness are shown in Table~\ref{tb:physical_values}, where three rows show our experiments under three weather/lighting conditions -- sunny, cloudy, and dusk weather. The values in this table reflect steering angle compared to the baseline steering without perturbation. Under each condition, the table shows the three aforementioned metrics for two adversarial settings (i.e., left-misleading (right-misleading) denoted by the ``Ad\_left'' (``Ad\_right'') column).
We observe that two adversarial perturbations both yield relatively large $Exp\_AAE$ (denoted by ``Exp'' in the table) and $Test\_AAE$ (denoted by ``Test'') for all weather conditions. For instance, under the Sunny condition, \tech~yields a left mis-leading steering angle of 8.88 degree. 
In many cases, the $Test\_AAE$ value is only slightly smaller than $Exp\_AAE$, indicating \tech{}'s effectiveness in the physical world setting. Furthermore, the percentage of frames that have a mis-steering angle larger than the pre-defined threshold (i.e., the $M_1$ metric) is more than 19\% in most scenarios (4 out of 6) and can even reach up to 100\%. 
Overall, we detected 268 frames out of the total 900 frames that exhibit a mis-steering angle larger than the threshold, implying dangerous driving behaviors within these frames.

\begin{figure}[t!]
\centering
\vspace{4mm}
\includegraphics[width=3.2in]{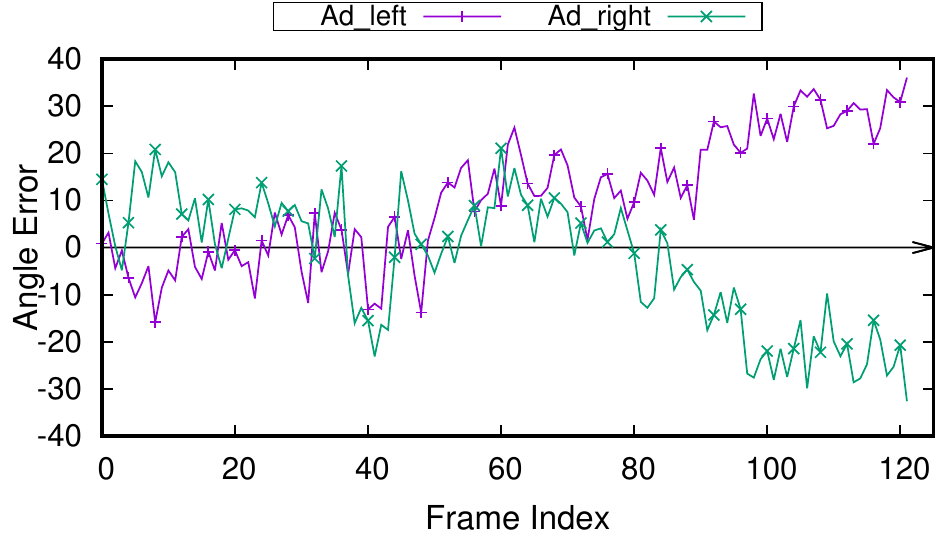}
\caption{Per-frame steering angle error.}
\vspace{4mm}
\label{fig:phy_timeline}
\end{figure}

To better interpret the results, we also report the per-frame steering angle for the physical tests under the Sunny condition in Fig.~\ref{fig:phy_timeline} (due to space constraints, we omit the other two scenarios which show similar result trends), where the x-axis represents the frame index, and the y-axis represents the steering angle. We note again that a positive (negative) steering angle value indicates a left (right) steering. 
The two curves indicated in this figure represent the testings of applying two adversarial billboards (left-misleading and right-misleading).
We observe that, the left-misleading test successfully causes the angle errors larger than zero, and the right-misleading test successfully causes the angle errors smaller than zero. This trend become mores observable at later frames, since the billboard becomes larger in each frame.

\section{Conclusion}
In this paper, we propose \tech, a systematic physical-world testing of autonomous driving systems. \tech~develops Robust Joint Optimization to systematically generate adversarial perturbation that can be patched on roadside billboards both digitally and physically to consistently cause mis-steering in a scene of multiple frames with different viewing distances and angles. Extensive experiment results demonstrate the efficacy of \tech~in testing various steering models in various digital and physical-world scenarios. Furthermore, the basic \tech approach can be directly generalized to a variety of other physical entities/surfaces besides billboards along the curbside, e.g., a graffiti painted on a wall.

\IEEEpeerreviewmaketitle


\bibliographystyle{IEEEtran}
\bibliography{dnn-test}

\end{document}